\documentclass[10pt,journal,compsoc]{IEEEtran}

\usepackage{epsfig}
\usepackage{graphicx}
\usepackage{amsmath}
\usepackage{amssymb}
\usepackage{mathtools}
\usepackage{bbm}
\usepackage{makecell, verbatim, sidecap}

\usepackage[nocompress]{cite}

\usepackage{balance}

\usepackage[breaklinks=true,bookmarks=false]{hyperref}

\usepackage{times}
\usepackage{epsfig}
\usepackage{graphicx}
\usepackage{amsmath}
\usepackage{amssymb}
\usepackage{bm}
\usepackage{algorithm}
\usepackage{algorithmic}
\usepackage{multirow}
\usepackage{eqparbox}
\usepackage{color}
\usepackage{float}
\usepackage{makecell}

\usepackage{color}
\usepackage[dvipsnames]{xcolor}
\usepackage{hhline}

\usepackage{multirow}
\usepackage{eqparbox}
\usepackage{makecell}

\usepackage{booktabs}
\usepackage{verbatim}
\newcommand{\green}[1] { \color[rgb]{0.13, 0.55, 0.13} {  $\!$ #1}}

\renewcommand{\texttt}[1]{${{\tt #1}}$}

\usepackage{xspace}

\renewcommand{\paragraph}{\textbf}

\begin{document}

    \title{Self-Supervised 3D Scene Flow Estimation and Motion Prediction using Local Rigidity Prior}

    \author{
         Ruibo Li,
         Chi Zhang,
        Zhe Wang,
        Chunhua Shen,
        Guosheng Lin
        \thanks{
            G. Lin is the
            corresponding author  (e-mail:  gslin@ntu.edu.sg).

        }
    }

    \IEEEtitleabstractindextext{

        \begin{abstract}
        	
In this article, we investigate self-supervised 3D scene flow estimation and class-agnostic motion prediction on point clouds. 
A realistic scene can be well modeled as a collection of rigidly moving parts, therefore its scene flow can be represented as a combination of the rigid motion of these individual parts. 
Building upon this observation, we propose to generate pseudo scene flow labels for self-supervised learning through piecewise rigid motion estimation, in which the source point cloud is decomposed into local regions and each region is treated as rigid. 
By rigidly aligning each region with its potential counterpart in the target point cloud, we obtain a region-specific rigid transformation to generate its pseudo flow labels. 
To mitigate the impact of potential outliers on label generation, when solving the rigid registration for each region, we alternately perform three steps: establishing point correspondences, measuring the confidence for the correspondences, and updating the rigid transformation based on the correspondences and their confidence.
As a result, confident correspondences will dominate label generation and a validity mask will be derived for the generated pseudo labels.
By using the pseudo labels together with their validity mask for supervision,  models can be trained in a self-supervised manner.
Extensive experiments on FlyingThings3D and KITTI datasets demonstrate that our method achieves new state-of-the-art performance in self-supervised scene flow learning, without any ground truth scene flow for supervision, even performing better than some supervised counterparts. 
Additionally, our method is further extended to class-agnostic motion prediction and significantly outperforms previous state-of-the-art self-supervised methods on nuScenes dataset.
  
 \end{abstract}

    }

    \maketitle

\begin{figure*}[]
	\centering
	\includegraphics[height=9.8cm]{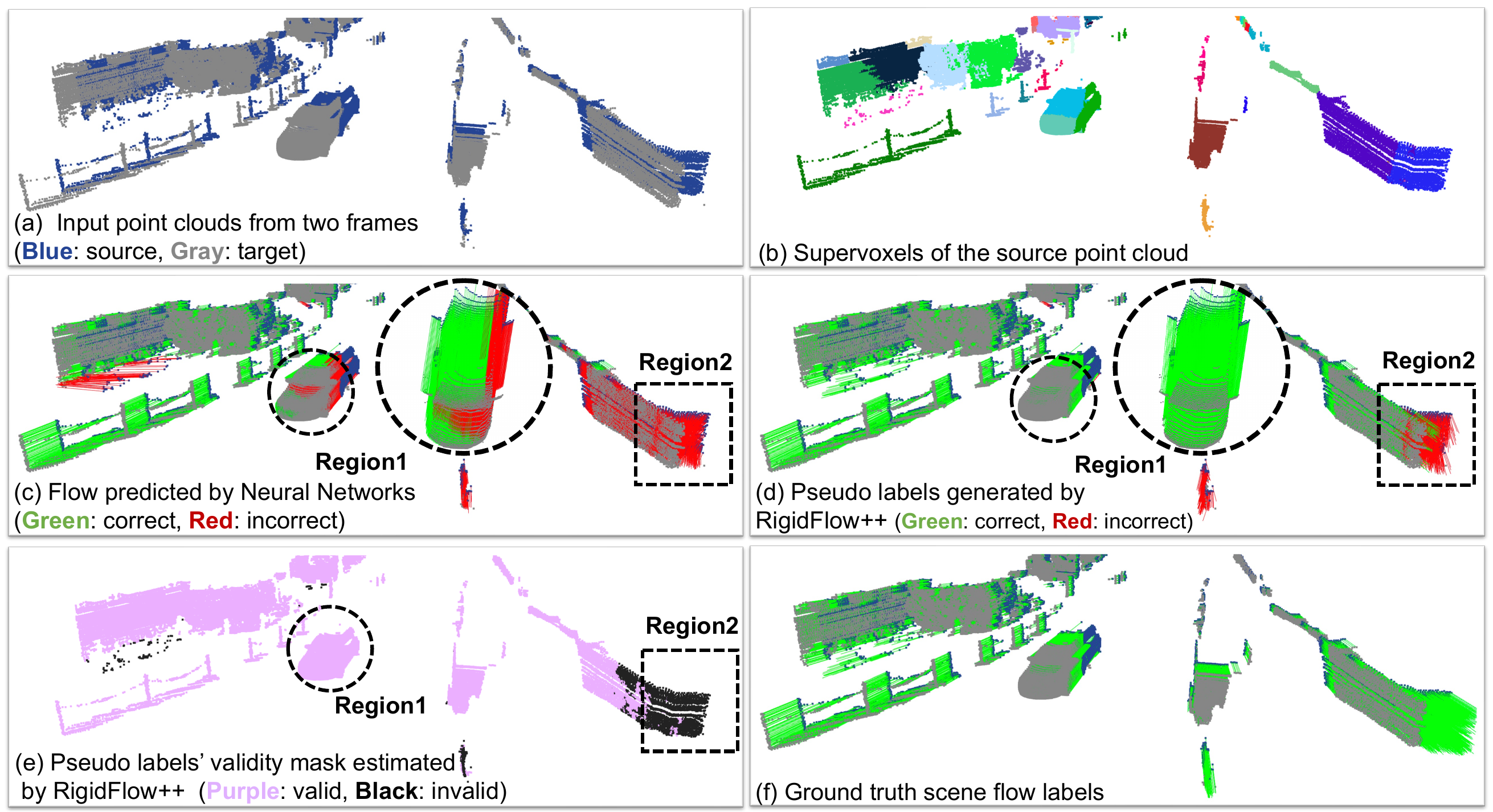}
		\vspace{-2mm}
	\caption{
	Illustration of pseudo scene flow labels and their validity mask generated by our proposed method, RigidFlow++. 
	(a) Input point clouds from two consecutive frames. 
	(b) Supervoxels of the source point cloud.
(c) Scene flow predictions from neural networks.
(d) Pseudo scene flow labels generated by RigidFlow++.
{\color{Green}\textbf{Green line}}  represents the {\color{Green}\textbf{correct}} flow or pseudo label with absolute error less than 0.1$m$ or relative error less than 10\%. {\color{Red}\textbf{Red line}} represents the {\color{Red}\textbf{incorrect}} flow or pseudo label.
(e) Binary validity mask for these pseudo labels estimated by RigidFlow++.
(f) Ground truth scene flow labels.
In Region 1, our generated pseudo labels are more accurate than the predicted flow, which allows the pseudo labels to serve as supervision.
In Region 2, the inaccurate pseudo labels are estimated to be invalid.
During training,  these invalid pseudo labels will be filtered out, and the valid ones will dominate the self-supervised training of neural networks.
}
	\label{pic1}	
\end{figure*}

\section{Introduction}
\label{sec:intro}

Scene flow estimation~\cite{vedula1999three}  aims to generate a 3D motion field of a dynamic scene.
As a fundamental representation of dynamics, scene flow can be applied in various tasks, such as motion segmentation~\cite{baur2021slim}, 3d object detection~\cite{erccelik20223d}, and point cloud accumulation~\cite{huang2022dynamic}, as well as multiple downstream applications including robotics and autonomous driving~\cite{menze2015object,najibi2022motion}.
In recent years, with the widespread application of 3D sensors and the rise of deep learning techniques for point cloud processing, learning scene flow directly from 3D point clouds has attracted increasing research attention.

However, the ground truth of scene flow is difficult to collect
~\cite{menze2015object}, which makes the supervised learning approaches suffer from a shortage of real-world training samples. 
While synthetic data, such as the FlyingThings3D dataset~\cite{mayer2016large}, can mitigate the need for expensive real-world scene flow data in supervised approaches~\cite{liu2019flownet3d,gu2019hplflownet,puy2020flot,wu2020pointpwc,kittenplon2021flowstep3d,li2021hcrf,wei2021pv}, the domain gap between synthetic and realistic data may lead to poor performance of models trained on synthetic data when applied to real-world scenes.
Apart from using synthetic scene flow data for training, some works~\cite{gojcic2021weakly,dong2022exploiting} propose to train models on realistic data in a weakly supervised manner.
Despite alleviating the reliance on scene flow annotations, these weakly supervised approaches still necessitate dense foreground/background annotations and ego-motion information for supervision.
In contrast to fully supervised approaches that utilize synthetic data and weakly supervised approaches that rely on dense foreground labels, our work studies self-supervised scene flow learning, where models can be trained on realistic data without using any manually annotated ground truth.

Scene flow describes the temporal connection between two consecutive point clouds. 
To enable deep network training under the self-supervised setting, in most previous approaches~\cite{mittal2020just,tishchenko2020self,kittenplon2021flowstep3d, baur2021slim, wu2020pointpwc,li2021self, pontes2020scene,gu2022rcp,he2022self,shen2023self}, models estimate scene flow between two point clouds, and then the estimated scene flow is used to warp the source point cloud to match the target one.
The main supervision signal is obtained by minimizing the discrepancy between the warped point cloud and the target point cloud, that is, by minimizing the distance between corresponding points in the two point clouds.
Specifically, when performing point matching to establish point correspondences, \cite{mittal2020just,kittenplon2021flowstep3d,baur2021slim,wu2020pointpwc} leverage nearest neighbor search, \cite{he2022self} follows Cauchy-Schwarz divergence, and \cite{li2021self} utilizes optimal transport. 
Although achieving promising performance, the point matching strategies employed in these approaches tend to neglect the potential structured motion of points.
Consequently, this oversight will result in inconsistent point correspondences that violate the constraints on the local rigidity of motion, thereby generating inaccurate supervision signals.

For a real-world scene, most structures in this scene are rigid or almost so~\cite{man1982computational}.
This allows us to decompose a non-rigid scene into a collection of rigidly moving parts, such that the entire scene flow can be approximated by estimating the rigid motion of individual parts.
Inspired by this observation,  in this work, we propose to generate pseudo scene flow labels via piecewise rigid motion estimations and use these pseudo labels as supervision signals for self-supervised learning.

To achieve this goal,  an over-segmentation approach is employed to decompose the source point cloud into supervoxels (Fig.~\ref{pic1}(b)), and these supervoxels are treated as rigid during the pseudo label generation.  
By solving an independent rigid registration for each supervoxel, we find a rigid transformation that rigidly aligns this supervoxel with its potential counterpart in the target point cloud. 
Based on the rigid transformation estimate, we generate the rigid flow for each supervoxel, thereby yielding locally rigid pseudo scene flow labels for the entire source point cloud.

To solve the piecewise rigid registration, iterative closest point (ICP)~\cite{besl1992method} is feasible.
This algorithm adopts an iterative procedure comprising two alternating steps: 
(1) estimating the rigid transformation by solving a least-squares problem according to the point correspondences;  
(2) warping points by the estimated rigid transformation and then updating the correspondences to their closest matches.
Nevertheless, the ICP algorithm is susceptible to the influence of outliers~\cite{yew2020rpm,gojcic2020learning}.
Unfortunately, due to the occlusion, sparsity, and noise of dynamic point clouds, outliers are widespread in the context of self-supervised 3D scene flow learning.
To tackle this issue, we introduce a confidence reweighting mechanism and estimate the rigid transformation by solving a weighted least-squares problem.
Specifically, we measure the confidence of each point correspondence and subsequently assign lower weights to unconfident correspondences.
In 3D scene flow learning, for a confident point correspondence, the forward flow of the source point should be the inverse of the backward flow of the target point, i.e., the constraint of forward-backward consistency, and the warped source point should be close to the target point, i.e., the constraint of spatial proximity.
Based on this insight, we assign lower weights to correspondences that deviate from these constraints to mitigate the impact of potentially erroneous correspondences on pseudo label generation.
After generating the final pseudo scene flow labels, we also produce a validity mask to indicate the validity of the pseudo labels according to the two constraints. 
By applying the validity mask together with the pseudo labels to loss functions, the invalid pseudo labels will be filtered out and the valid ones will dominate the self-supervised training of scene flow models.
Fig.~\ref{pic1} provides an example of pseudo scene flow labels and their validity mask. 

Class-agnostic motion prediction aims to generate the future positions of objects based on previous observations, which holds significance for path planning and navigation.
Given a series of point clouds from past frames, some works~\cite{wu2020motionnet,luo2021self,wang2022sti,li2023weakly} propose to convert the past point clouds into bird’s eye view (BEV) maps and output a motion vector of each cell in the current BEV map to indicate its displacement from the current to the future.
Treating the BEV-based motion field as scene flow, we generate pseudo BEV-based motion labels via our self-supervised scene flow method and use these pseudo labels to train motion prediction models.
By this means, we extend the applicability of our method to self-supervised class-agnostic motion prediction.

The main contributions of this paper are listed as follows:
\begin{itemize}
	\item  
	We present a new self-supervised scene flow learning approach (RigidFlow++) that solves the pseudo scene flow label generation as a piecewise rigid motion estimation task.
	
	\item  
	By decomposing the source point cloud into a set of local regions, we propose a confidence-aware piecewise pseudo label generation module that alternately estimates point correspondences, confidence weights, and rigid transformations to generate reliable rigid pseudo flow labels and their validity mask for each local region.
	
	\item  
	Our proposed RigidFlow++ achieves state-of-the-art performance in self-supervised scene flow learning, without any ground truth scene flow for supervision, even outperforming some supervised counterparts.
	
	\item  
	We further extend our proposed RigidFlow++ to the task of self-supervised class-agnostic motion prediction and achieve superior performance compared to the previous state-of-the-art. 
\end{itemize}

A preliminary version of this work was published in~\cite{li2022rigidflow}.
We have extended the conference version from several aspects:
\textbf{(i)} We improve our original method (RigidFlow) by introducing a confidence reweighting mechanism to suppress potential outliers, resulting in more robust pseudo label generation.
In addition, we further reason about the validity of pseudo labels, enabling more effective self-supervised training. 
The experimental results indicate that these enhancements lead to substantial improvements, particularly in scenes with occlusions.
\textbf{(ii)}  We conduct more extensive experiments to validate the effectiveness of our self-supervised method, including comparisons with recent test-time optimization-based methods and an exploration of the generalization capability of our method across diverse scene flow models.
\textbf{(iii)} We further extend our method to self-supervised class-agnostic motion prediction and achieve state-of-the-art performance. 
\textbf{(iv)} We include more comprehensive literature on the state-of-the-art scene flow estimation methods and introduce some related works on class-agnostic motion prediction. 
The code and models of the preliminary version are in this link\footnote{https://github.com/L1bra1/RigidFlow}. And the code and models for this new version will be made publicly available.

\section{Related Work}
\noindent\textbf{Scene flow estimation on images.}\quad 
Scene flow~\cite{vedula1999three} refers to a 3D motion field that describes the movement of objects in a dynamic scene, and it can be estimated via different types of data.
 Nowadays, the local rigidity assumption has been widely employed in numerous advanced approaches~\cite{teed2021raft, jaimez2015motion, kumar2017monocular, ma2019deep, menze2015object, vogel20113d, vogel2013piecewise, vogel20153d, hornacek2014sphereflow, liu2019unsupervised, behl2017bounding, lv2018learning, jiao2021effiscene} to facilitate scene flow estimation from RGB images or  RGB-D images. 
For example, UnRigidFlow~\cite{liu2019unsupervised} and EffiScene~\cite{jiao2021effiscene} address unsupervised scene flow estimation from images by jointly learning rigidity masks to constrain scene flow predictions.
Different from these methods that use well-organized 2D images as input and employ photometric error as the major loss function, our work focuses on scene flow estimation from irregular and sparse 3D point clouds and explores the application of the local rigidity to pseudo scene flow label generation, thus self-supervised scene flow learning can be achieved by any supervised loss functions with our generated pseudo labels.\\

\noindent\textbf{Supervised scene flow estimation on point clouds.}\quad 
The development of 3D sensors has led to a growing interest in scene flow estimation from point clouds.
Various approaches~\cite{liu2019flownet3d, behl2019pointflownet, liu2019meteornet, gu2019hplflownet, puy2020flot, wei2021pv, li2021hcrf, gojcic2021weakly,li2022sctn,he2022learning,ding2022fh,li2022rppformer,wang2022matters,cheng2022bi,wu2023pointconvformer} have been proposed to achieve scene flow estimation in a fully supervised manner.
Specifically, 3DFlow~\cite{wang2022matters} estimates scene flow by establishing all-to-all flow embedding.
And BiFlowNet~\cite{cheng2022bi} estimates scene flow by capturing multi-scale bidirectional flow correlation.
Recently, the two models have demonstrated strong performance in scene flow learning.
In this work, we evaluate the generalization capability of our self-supervised method across different models by applying it to 3DFlow and BiFlowNet.
Apart from fully supervised scene flow learning, some works~\cite{gojcic2021weakly,dong2022exploiting} study scene flow estimation in a weakly supervised manner.
Despite alleviating the reliance on scene flow annotations, these approaches still necessitate dense foreground/background annotations as weak supervision for training.
In contrast to fully and weakly supervised approaches, our work studies self-supervised scene flow learning without using any manually annotated ground truth.

Particularly, the local rigidity has also been applied in some of the supervised models~\cite{behl2019pointflownet, gojcic2021weakly, dong2022exploiting, li2021hcrf}.
PointFlowNet~\cite{behl2019pointflownet}, Rigid3DSceneFlow~\cite{gojcic2021weakly}, and LiDARSceneFlow~\cite{dong2022exploiting} propose to estimate rigid motion for each 3D object directly.
HCRF-Flow~\cite{li2021hcrf} employs local rigidity to refine scene flow predictions.
Unlike these supervised approaches that learn to refine or constrain predicted flow by the local rigidity, we explore how to produce pseudo labels with the guidance of the local rigidity assumption to achieve self-supervised scene flow learning.\\

\noindent\textbf{Self-supervised scene flow estimation on point clouds.}\quad 
In self-supervised scene flow learning, given two consecutive point clouds without ground truth, most previous methods~\cite{mittal2020just,tishchenko2020self,kittenplon2021flowstep3d, baur2021slim, wu2020pointpwc,pontes2020scene,li2021self,gu2022rcp,he2022self,shen2023self,ouyang2021occlusion} estimate scene flow by models and warp the source point cloud by the estimated scene flow to match the target one.
The self-training of models is performed by minimizing the distance between corresponding points in the two point clouds.
Specifically, \cite{mittal2020just, baur2021slim, tishchenko2020self} adopt a nearest neighbor loss and \cite{wu2020pointpwc, kittenplon2021flowstep3d, pontes2020scene,gu2022rcp,shen2023self,ouyang2021occlusion} adopt a Chamfer loss for self-training. 
The two loss functions build point correspondences by nearest neighbor search. 
Additionally, Self-Point-Flow~\cite{li2021self} establishes point correspondences via solving an optimal transport.
And PDF-Flow~\cite{he2022self} represents point clouds by probability density functions to build soft point correspondences.
However, the point matching strategies employed in these approaches tend to ignore the potential structured motion of points, leading to inaccurate supervision signals.

Motivated by the local rigidity assumption, we propose to generate pseudo labels by piecewise rigid motion estimation.
By explicitly enforcing region-wise rigid alignments between the source and target point clouds, our method generates locally rigid pseudo scene flow labels as supervision for self-training.
Although the rigidity of motion has been used in some recent self-supervised works~\cite{pontes2020scene,baur2021slim,shen2023self}, the clue of rigidity is limited to smooth or constrain the flow predictions from models and the loss functions adopted in these works are still based on the nearest neighbor loss or the Chamfer loss.
Therefore, the potential of local rigidity in improving pseudo label generation to provide more accurate supervision signals is far from being explored, and it is the focus of our work.
Furthermore, our preliminary method~\cite{li2022rigidflow}, RigidFlow, may fail to handle outliers caused by the occlusion, sparsity, and noise of dynamic point clouds.
To address this issue, our improved method, RigidFlow++, leverages the constraints of forward-backward consistency and spatial proximity to suppress potential outliers in pseudo label generation and reason about the validity of the generated pseudo labels for more efficient self-supervised learning. 
Compared to RigidFlow, RigidFlow++ achieves substantial improvements, particularly in scenes with occlusions.\\

\noindent\textbf{Test-time optimization-based scene flow estimation on point clouds.}\quad 
When ground truth data is unavailable, apart from self-supervised scene flow learning, some works~\cite{pontes2020scene,li2021neural,najibi2022motion,lang2022scoop,deng2023rsf} design offline optimization approaches to estimate scene flow at test time.
Specifically, NSFP~\cite{li2021neural} optimizes scene flow for each scene separately using a neural network as the regularizer.
And SCOOP~\cite{lang2022scoop} estimates scene flow by combining a self-supervised neural network and a test-time optimization-based refinement.
While these approaches achieve good performance, they tend to be highly time-consuming in testing due to the extensive iterations required for optimization.
Different from these methods, our self-supervised method is only performed in the training of models and does not introduce any extra runtime to the models in testing.
This characteristic renders our self-supervised method suitable for time-sensitive and low-power applications. \\

\noindent\textbf{Class-agnostic motion prediction.}\quad 
Motion prediction aims to estimate the future positions of objects based on past observations.
Given consecutive point clouds from past frames, some works~\cite{wu2020motionnet, luo2021self, wang2022sti, li2023weakly,filatov2020any,lee2020pillarflow,schreiber2021dynamic} propose to convert the point clouds into bird’s eye view (BEV) maps and estimate a future motion field from these BEV maps.
MotionNet~\cite{wu2020motionnet} learns to simultaneously estimate both semantic information and future motion in a supervised manner.
And PillarMotion~\cite{luo2021self} presents a self-supervised training strategy that employs Chamfer distance as the loss function and leverages 2D optical flow of RGB images to provide motion regularization.
Although PillarMotion has achieved good results, RGB images and optical flow estimation networks may not be available in some scenarios, limiting the application of this method.

Regarding the future motion field as scene flow, we generate pseudo motion labels via our self-supervised scene flow method and use these pseudo labels to train a motion prediction network built on MotionNet in a self-supervised manner.
Compared with PillarMotion, our method achieves superior performance without using any images or optical flow networks.

\section{Preliminaries: Rigid registration and ICP}
As a crucial task in computer vision, point cloud registration has been well studied
in the literature~\cite{besl1992method, fujiwara2011locally, rusinkiewicz2001efficient, segal2009generalized, wang2019prnet, wang2019deep}.
Given two point clouds, ${\bm X} = \{{\bm x}_i \in  \mathbb{R}^3 \}_{i=1}^{N_x}$ and ${\bm Y} = \{{\bm y}_i \in  \mathbb{R}^3 \}_{i=1}^{N_y}$, point cloud rigid registration aims
to predict a rigid transformation that aligns  $\bm X$ to $\bm Y$.
The rigid transformation can be written as $[\bm{R}_{ XY}, {\bm t}_{ XY}]$, where the rotation matrix ${\bm R}_{ XY} \in {\rm SO}(3)$ and the translation vector ${\bm t}_{XY} \in \mathbb{R} ^3$.
The objective function of point cloud registration can be expressed as:
\begin{equation}\label{Reg}
{{E}}({\bm R}_{XY}, {\bm t}_{XY}; {\bm X}, {\bm Y})= \sum_{i = 1}^{n}{\| {\bm R}_{XY} \bm{x}_i + {\bm t}_{XY} - {{\bm y}_{ m(i)} }  \|_2^2},
\end{equation}
where $\bm m$ is the point mapping from points in $\bm X$ to their corresponding points in $\bm Y$.

Since the point mapping $\bm m $ is unknown, the iterative closest point (ICP)~\cite{besl1992method}  is widely employed to address this problem by alternating between estimating the rigid transformation and finding the point mapping.
In each iteration, based on the previous point mapping estimate, the current rigid transformation is updated by solving the least-squares problem in Eq.~(\ref{Reg}).
And then, by warping  $\bm X$ with the current rigid transformation estimate, the point mapping of each point in $\bm X$ is updated to its closest match in  another point cloud:
\begin{equation}\label{Reg_m}
{m}_i=  \arg \min_j \| {\bm R}_{XY} \bm{x}_i + {\bm t}_{XY} - {{\bm y}_j }  \|_2^2.
\end{equation}
Although the ICP is efficient, the performance depends heavily on the initialization of rigid transformation and point matching.

\begin{figure*}[tb]
	\centering 
	\includegraphics[height=11cm]{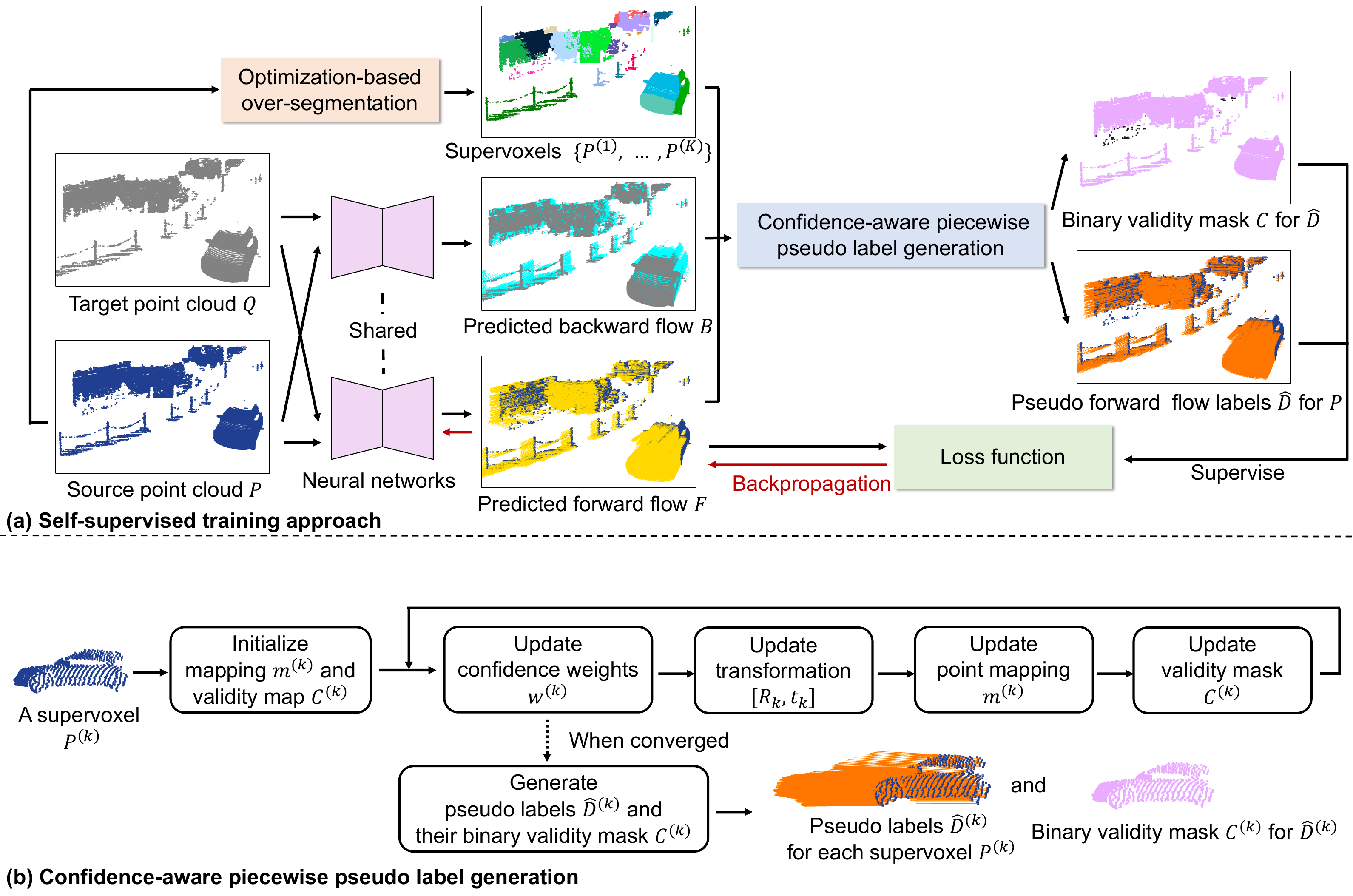}
	\vspace{-2mm}
	\caption{(a) Overview of our self-supervised training approach. In our approach, we first employ an optimization-based over-segmentation method~\cite{lin2018toward} to split the source point cloud  ${\bm P}$ into a set of supervoxels $\{  {\bm P}^{(1)}, ...,  {\bm P}^{(K)}  \}$.
		And then, we use the confidence-aware piecewise pseudo label generation module to generate pseudo forward flow labels  ${\bm {\widehat D}}^{(k)} $  and their binary validity mask  ${ C}^{(k)}$  for each supervoxel ${\bm P}^{(k)}$ by estimating the rigid transformation.
		Finally, the self-supervised training of neural networks can be achieved using the pseudo forward flow labels  ${\bm {\widehat D}}$ and the binary validity mask ${ C}$ as supervision. 
		(b) Illustration of confidence-aware piecewise pseudo label generation module. For each supervoxel ${\bm P}^{(k)}$, we alternately estimate point mapping ${\bm m^{(k)}}$, confidence weights ${ \bm w^{(k)}}$, and rigid transformation $[\bm{R}_k, \bm{t}_k]$ to generate pseudo labels  ${\bm {\widehat D}}^{(k)} $ and their binary validity mask ${ C}^{(k)}$. }
	\label{fig_pipeline}	
\end{figure*}

\section{Method}
Given a temporal sequence of point clouds,  ${\bm P} = \{{\bm p}_i \in  \mathbb{R}^3 \}_{i=1}^{N}$ at source frame $t$ and  ${\bm Q} = \{{\bm q}_i \in  \mathbb{R}^3 \}_{i=1}^{N}$ at target frame $t+1$, scene flow estimation aims to produce the forward 3D motion field ${\bm F} = \{{\bm f}_i \in  \mathbb{R}^3 \}_{i=1}^{N}$ in a scene.
In this paper, we target self-supervised point cloud scene flow estimation, where no ground truth scene flow labels $\bm D$ are provided.
To enable network training without ground truth, we focus on effective pseudo scene flow label generation.
The overview of our self-supervised training approach is illustrated in Fig.~\ref{fig_pipeline}~(a).
In our framework, we first employ an optimization-based over-segmentation method~\cite{lin2018toward} to split ${\bm P}$ into a set of supervoxels.
And then, we use our designed confidence-aware piecewise pseudo label generation module to produce pseudo forward scene flow labels  ${\bm {\widehat D}}^{(k)} $ and their binary validity mask ${ C}^{(k)}$ for each supervoxel ${\bm P}^{(k)}$.
Specifically, we predict the backward scene flow $\bm B$ as auxiliary information and measure the confidence by the consistency of forward and backward flows and the spatial proximity of warped source point and target point.
Finally, using the entire pseudo forward flow labels ${\bm {\widehat D}} $ and their binary validity mask ${ C}$ as supervision, the self-supervised training of neural networks can be achieved by minimizing the discrepancy between ${\bm {\widehat D}} $ and ${\bm {F}} $.

In this section, we first introduce how to produce pseudo scene flow for a real-world scene by robust piecewise rigid motion estimation  (Sec.~\ref{Robust_Reg}).
After that, we present the details of our confidence-aware  piecewise pseudo label generation module (Sec.~\ref{Label_gen}).
Finally, we describe how to use the generated pseudo labels and their validity mask to achieve self-supervised training (Sec.~\ref{Loss_func}).

\subsection{Generating Pseudo Labels by Robust Piecewise Rigid Motion Estimation}\label{Robust_Reg}
Scene flow represents the 3D motion field of objects in a scene.
If the scene only contains a single rigidly moving object, the scene flow from $\bm P_{rigid}$ to $\bm Q_{rigid}$ follows the rigid transformation $[{\bm R}_{PQ}, {\bm t}_{ PQ}]$ between the two point clouds:
\begin{equation}\label{basic_rigid}
{\bm D} = \bm{R}_{PQ} {\bm P_{rigid}} + {\bm t}_{PQ} - {\bm P_{rigid}}.
\end{equation}
Therefore, for a rigidly moving object,  when the ground truth scene flow is unavailable, we can produce the scene flow by finding its optimal rigid transformation between $\bm P_{rigid}$ and $\bm Q_{rigid}$.

For a complex real-world scene, although it is not rigid, most of the structures in this scene are rigid or almost rigid, which makes it possible to approximate a non-rigid scene into a set of rigidly moving regions. 
Therefore, we can estimate the flow of each rigidly moving region by finding its optimal rigid motion, thereby generating scene flow for the entire non-rigid scene.
In other words, we can perform a piecewise rigid motion estimation to generate scene flow that can serve as pseudo labels for self-supervised learning.

Decomposing the point cloud $\bm P$ into $K$ rigid regions $\{  {\bm P}^{(1)},  {\bm P}^{(2)}, ...,  {\bm P}^{(K)}  \}$, the piecewise rigid motion estimation from $\bm P$ to $\bm Q$ for the region ${\bm P}^{(k)}$ can be considered as an independent rigid body  registration from ${\bm P}^{(k)}$ to ${\bm Q}$:
\begin{equation}\label{P1}
	\begin{aligned}
	[\bm{R}^*_k, \bm{t}^*_k] = &\mathop{\arg\min}_{[\bm{R}_k, \bm{t}_k]}{\sum_{i = 1}^{N_k}{\| {\bm R}_{k} {\bm p}^{(k)}_i + {\bm t}_{k} - {{\bm q}_{ m^{(k)}_i} }  \|_2^2}} \\
	&\text{s.t.} \quad   {\bm R}_{k} \in SO(3),
	\end{aligned}
\end{equation}
where $N_k$ is the point number in the rigid region ${\bm P}^{(k)}$, ${ m^{(k)}_i}$ is the mapping from the $i$-th point in ${\bm P}^{(k)}$ to its correspondence in $\bm Q$, and $[\bm{R}^*_k, \bm{t}^*_k]$ is the optimal rigid transformation for ${\bm P}^{(k)}$.
When applying the ICP algorithm~\cite{besl1992method} to this registration problem, we solve the rigid transformation and search the point mappings alternatively.
By fixing the current point mappings, the optimal rigid transformation in each iteration can be retrieved by solving the least-squares problem in Eq.~(\ref{P1}).  

However, the least-squares solution of registration is not robust~\cite{gojcic2020learning}, which may produce inaccurate transformation results when the ratio of outliers is high.
Unfortunately, due to the view-changes, occlusions, sparsity, and noise of dynamic point clouds, outliers are widespread in the context of self-supervised 3D scene flow learning.

To address this issue, we further propose to achieve pseudo label generation with a robust piecewise rigid motion estimation. 
For each rigid region ${\bm P}^{(k)}$,  the robust piecewise rigid motion estimation can be formulated as a weighted registration from ${\bm P}^{(k)}$ to ${\bm Q}$:
\begin{equation}\label{P2}
\begin{aligned}
[\bm{R}^*_k, \bm{t}^*_k] = & \mathop{\arg\min}_{[\bm{R}_k, \bm{t}_k]}{\sum_{i = 1}^{N_k}
	{  {w}^{(k)}_i \| {\bm R}_{k} {\bm p}^{(k)}_i + {\bm t}_{k} - {{\bm q}_{ m^{(k)}_i} }  \|_2^2}}
 \\
&\text{s.t.} \quad   {\bm R}_{k} \in SO(3),
\end{aligned}
\end{equation}
where ${w}^{(k)}_i \in [0,1]$ is a weight to measure the confidence score of the mapping from point  $ {\bm p}^{(k)}_i $ to its correspondence $ {{\bm q}_{ m^{(k)}_i} } $. 
Compared to the original registration,  Eq.~(\ref{P1}), that assigns the same weight to each point mapping, the robust registration, Eq.~(\ref{P2}), will assign confident point mappings higher weights, so that the confident point mappings will dominate the solution, thereby  producing more accurate rigid transformations for pseudo label generation.

By solving the robust registration, we obtain the optimal rigid transformation $[\bm{R}^*_k, \bm{t}^*_k]$ for ${\bm P}^{(k)}$.
Following Eq.~(\ref{basic_rigid}), the pseudo rigid scene flow estimate ${\bm {\widehat D}}^{(k)}$ for this region can be computed by:
\begin{equation}\label{region_rigid}
{\bm {\widehat D}}^{(k)} = \bm{R}^*_k {\bm P}^{(k)} + \bm{t}^*_k - {\bm P}^{(k)}.
\end{equation}
Combining the pseudo rigid scene flow estimates for all $K$ rigid regions $\{ {\bm {\widehat D}}^{(1)}, {\bm {\widehat D}}^{(2)}, ...,  {\bm {\widehat D}}^{(K)}  \}$, we obtain the final pseudo rigid scene flow estimate ${\bm {\widehat D}}$ as pseudo scene flow labels for self-supervised training.

\subsection{Confidence-aware  Piecewise Pseudo Label Generation Module}\label{Label_gen}
In order to convert pseudo label generation into a robust piecewise rigid motion estimation, we first employ an over-segmentation method~\cite{lin2018toward} to split the source point cloud $\bm P$ into supervoxels and treat these supervoxels as rigid moving regions. 
After obtaining the supervoxels, we  will generate pseudo scene flow labels for each supervoxel ${\bm P}^{(k)}$  by solving a weighted registration from ${\bm P}^{(k)}$ to ${\bm Q}$, where different point mappings are assigned with different confidence weights to suppress potential outliers.

In this module, we measure the confidence weights from two aspects.
Firstly, as presented in Eq.~(\ref{Reg_m}), the mapping of each point is established by finding the closest point to its warped point in another point cloud.
If the point mapping is reliable, the warped point should be
close to the closest match, i.e., the constraint of spatial proximity.
Therefore, we consider the point mapping with a large distance between the warped point and its closest match as invalid, and set its confidence weight to zero.
Secondly, inspired by the forward-backward consistency in 2D optical flow estimation~\cite{sundaram2010dense,meister2018unflow}, we also use the consistency of bidirectional 3D scene flow derived from the point mapping to measure its validity and confidence. 
For a reliable point mapping from ${\bm p}_i $ in ${\bm P}$ to its correspondence ${\bm q}_i$ in $\bm Q$, the forward flow of ${\bm p}_i $ should be the inverse of the backward flow of ${\bm q}_i$.
Therefore, we consider the point mapping with a large mismatch between the forward flow and the backward flow as invalid, and set the confidence weight to zero.
Subsequently, among these valid point mappings, the ones with more consistent bidirectional flows will be considered more confident and assigned higher confidence weights, thus dominating the generation of pseudo labels.

Following the principle of ICP algorithm~\cite{besl1992method}, in this confidence-aware  pseudo label generation module, we propose to alternately estimate point mappings, confidence weights, and rigid transformations, thereby generating pseudo scene flow labels with the estimated rigid transformations.
An illustration of this module is presented in Fig.~\ref{fig_pipeline}~(b).
Next, we present the details of this module. 

\subsubsection{Initialization}
\textbf{Initializing point mapping by predicted flow.}\quad 
The performance of ICP relies greatly on the initialization of rigid transformation and point mapping.
When solving the registration from ${\bm P}^{(k)}$ to ${\bm Q}$, for each point ${\bm p}^{(k)}_i$ in ${\bm P}^{(k)}$, a straightforward way of initialization is to set its closest point in ${\bm Q}$ as the initial correspondence.
Inspired by~\cite{li2021self}, we establish the initial point mapping based on the predicted forward scene flow $\bm F$ from neural networks being trained.
Specifically, we warp the point ${\bm p}^{(k)}_i$ by its predicted forward flow ${\bm f}^{(k)}_i$, and then take the closest point to this warped point as the initial match:
\begin{equation}\label{m_initial}
{ m_{i, init}^{(k)}} =  \mathop{\arg\min}_j \|   {\bm p}^{(k)}_i  + {\bm f}^{(k)}_i - {{\bm q}_j }  \|_2^2.
\end{equation}
As the training progresses,  the accuracy of the predicted scene flow will be gradually improved, making  the closest search of the warped points more likely to find the correct matches and establish good initial point correspondences.

\textbf{Initializing validity mask.}
For each point ${\bm p}^{(k)}_i$, according to its initial point mapping $ m^{(k)}_{i, init}$, we find out this corresponding point  ${\bm q}_{ m^{(k)}_{i, init}} $ and the backward predicted flow $ {\bm b}_{ m^{(k)}_{i, init}} $.
The point mapping with a large mismatch between the forward flow and the reversed backward flow or a large distance between the warped point and its correspondence will be regarded as invalid.
Therefore, the validity for the point mapping of ${\bm p}^{(k)}_i$ can be written as:
\begin{equation}\label{cm_initial}\small
{ C^{(k)}_i}  =  \text{I}( \|  {\bm f}^{(k)}_i + {\bm b}_{ m^{(k)}_{i, init}}   \|_2<\beta_1)  \cdot \text{I}( \|  {\bm p}^{(k)}_i +  {\bm f}^{(k)}_i  - {\bm q}_{ m^{(k)}_{i, init}}   \|_2< \beta_2), 
\end{equation}
where ${ C^{(k)}_i} \in \{0, 1\}$, $\text{I}(\cdot )$ denotes the indicator function and $\beta_1, \beta_2$ are the threshold values.
On the right side of Eq.~(\ref{cm_initial}), the first term is to measure the mismatch of bidirectional flows, and the second term is to measure the distance between the warped point and the corresponding point.

\begin{algorithm}[t]
	\caption{Pseudo scene flow label generation by confidence-aware piecewise rigid motion estimation}
	\label{alg1}
	\hspace*{0.02in}{\bf Input:} 
	\hspace*{0.1in}Source point cloud, $\bm P$;\\
	\hspace*{0.53in}Target point cloud, $\bm Q$;\\
	\hspace*{0.53in}Predicted forward  flow from NNs being trained. $\bm F$;\\
	\hspace*{0.53in}Predicted backward flow from NNs being trained, $\bm B$;\\
	\hspace*{0.02in}{\bf Output:} 
	\hspace*{0.02in}Pseudo scene flow labels, ${\bm {\widehat D}}$;\\
	\hspace*{0.53in}Binary validity mask, $\bm C$;\\
	\hspace*{0.02in}{\bf Procedure:} 
	\begin{algorithmic}[1]
		\STATE {Split $\bm P$ into a set of supervoxels $\{  {\bm P}^{(1)}, ...,  {\bm P}^{(K)}  \}$;}\\
		\COMMENT{Oversegmentation}
		\FOR{$k = 1,...,K$}
		\STATE {${ m^{(k)}_{i, init}} \gets  \arg \min_j \|   {\bm p}^{(k)}_i  + {\bm {f}}^{(k)}_i - {{\bm q}_j }  \|_2^2$}\\
		\COMMENT{ Initializing by forward flow}
		\STATE {${ C^{(k)}_i}  \gets  \text{I}( \|  {\bm f}^{(k)}_i + {\bm b}_{ m^{(k)}_{i, init}}   \|_2<\beta_1)  \cdot \text{I}( \|  {\bm p}^{(k)}_i +  {\bm f}^{(k)}_i  - {\bm q}_{ m^{(k)}_{i, init}}   \|_2< \beta_2) $}\\
		\COMMENT{ Initializing validity mask}
		
		\WHILE{ not converged}
		\STATE {Find the matches  ${\bm Q}^{(k)} $  and their predicted backward flow ${\bm B}^{(k)}$ for ${\bm P}^{(k)} $ based on ${\bm m^{(k)}}$ }
		\STATE {${ w}^{(k)}_i \gets \exp(\frac{-  \| {\bm f}^{(k)}_i  + {\bm b}^{(k)}_i  \|_2^2}{2 \theta^2}) \cdot { C^{(k)}_i}$}\\
		\COMMENT{Updating confidence weights}
		\STATE {${\bm H}_k \gets \sum_{i = 1}^{N_k} w^{(k)}_i  ({\bm p}^{(k)}_i -  \overline{\bm p}^{(k)}) ({\bm q}^{(k)}_i -  \overline{\bm q}^{(k)})^\top$}
		\STATE {$ 	\bm{R}_k \gets {\bm{V}_k} \  \text{diag}(1,1, \text{det}({\bm{V}_k}{\bm{U}_k}^\top))  \ {\bm{U}_k}^\top $}
		\STATE {$\bm{t}_k \gets  - \bm{R}_k \overline{\bm p}^{(k)} + \overline{\bm q}^{(k)} $}\\
		\COMMENT{Updating rigid transformation}
		\STATE {$ { m^{(k)}_i} \gets  \arg \min_j \| \bm{R}_k  {\bm p}^{(k)}_i  + \bm{t}_k - {{\bm q}_j }  \|_2^2 $}\\
		\COMMENT{Updating point mapping}
		\STATE {${ C^{(k)}_i} \gets   \text{I}( \|  {\bm f}^{(k)}_i + {\bm b}_{ m^{(k)}_i}   \|_2<\beta_1)  \cdot \text{I}( \|  \bm{R}_k  {\bm p}^{(k)}_i  + \bm{t}_k - {\bm q}_{ m^{(k)}_i}   \|_2< \beta_2) $}\\
		\COMMENT{ Updating validity mask}
		\ENDWHILE
		\STATE {${\bm {\widehat D}}^{(k)} \gets \bm{R}^*_k {\bm P}^{(k)} + \bm{t}^*_k - {\bm P}^{(k)}$}\\
		\COMMENT{Generating pseudo labels}
		\ENDFOR
		\STATE {${\bm {\widehat D}} \gets  \{  {\bm {\widehat D}}^{(1)}, ...,  {\bm {\widehat D}}^{(K)}  \} $; ${ \bm C} \gets  \{  { C}^{(1)}, ...,  { C}^{(K)}  \} $}
	\end{algorithmic}
\end{algorithm}

\subsubsection{Updating}
\textbf{Updating confidence weights.}\quad 
For the points in supervoxel ${\bm P}^{(k)}$, we first select their matches ${\bm Q}^{(k)}$ from $\bm Q$ according to the previous point mapping estimate ${\bm m^{(k)}}$. And then we find the backward scene flow $\bm B^{(k)}$ of ${\bm Q}^{(k)}$.
For each point mapping ${ m^{(k)}_i}$, we take the consistency between ${\bm f}^{(k)}_i $ and ${\bm b}^{(k)}_i $ as a metric and use a Gaussian kernel to generate its confidence score:
\begin{equation}\label{coarseweights}
{\bar w}^{(k)}_i =\exp(\frac{-  \| {\bm f}^{(k)}_i  + {\bm b}^{(k)}_i  \|_2^2}{2 \theta^2}),
\end{equation}
where $\theta$ is the kernel’s bandwidth parameter.
After obtaining the confidence score for each point mapping, we generate its confidence weight by filtering out invalid ones using the binary validity mask ${ C^{(k)}_i}$: 
\begin{equation}\label{weights}
w^{(k)}_i = {\bar w}^{(k)}_i  \cdot { C^{(k)}_i}.   
\end{equation}

\textbf{Updating rigid transformation estimate.}\quad 
Based on the previous point mapping estimate, we update the rigid transformation for each supervoxel by solving the weighted least-squares problem shown in Eq.~(\ref{P2}) with the point mapping fixed. Specifically, following~\cite{gojcic2020learning,sorkine2017least,torr1997development}, we apply the singular value decomposition (SVD)  to it.

Given the points in supervoxel ${\bm P}^{(k)}$ and their matches ${\bm Q}^{(k)}$ in $\bm Q$, the weighted centroids of ${\bm P}^{(k)}$ and ${\bm Q}^{(k)}$ are defined as
\begin{equation}\label{pq_center}
\overline{\bm p}^{(k)} = \frac{\sum\nolimits_{i = 1}^{N_k}{ w^{(k)}_i {\bm p}^{(k)}_i }}{\sum\nolimits_{i = 1}^{N_k}{ w^{(k)}_i}}  ,  \quad
\overline{\bm q}^{(k)} = \frac{\sum\nolimits_{i = 1}^{N_k}{ w^{(k)}_i {\bm q}^{(k)}_i }}{\sum\nolimits_{i = 1}^{N_k}{ w^{(k)}_i}} .
\end{equation}

The weighted cross-covariance matrix for supervoxel ${\bm P}^{(k)}$ can be written as:
\begin{equation}\label{H}
{\bm H}_{k} = \sum_{i = 1}^{N_k} w^{(k)}_i  ({\bm p}^{(k)}_i -  \overline{\bm p}^{(k)}) ({\bm q}^{(k)}_i -  \overline{\bm q}^{(k)})^\top.
\end{equation}
Using SVD to decompose ${\bm H}_{k}$, we have  ${\bm H}_{k} = \bm{U}_{k} \bm{S}_{k} {\bm{V}_{k}} ^\top$. The rotation matrix for supervoxel ${\bm P}^{(k)}$ can be updated in closed-form as:
\begin{equation}\label{R_update}
\bm{R}_k  = 
{\bm{V}_k} \begin{bmatrix}
1 & 0 & 0 \\
0 & 1 & 0 \\
0 & 0 & \text{det}({\bm{V}_k}{\bm{U}_k}^\top)
\end{bmatrix}{\bm{U}_k}^\top,
\end{equation}
where $\text{det}(\cdot )$ denotes the determinant of a matrix.
And the translation vector can be updated by:
\begin{equation}\label{t_update}
\bm{t}_k   = - \bm{R}_k \overline{\bm p}^{(k)} + \overline{\bm q}^{(k)}.
\end{equation}

\textbf{Updating point mapping estimate.}\quad 
Warping the points in supervoxel ${\bm P}^{(k)}$ by the current rigid transformation estimate $[\bm{R}_k, \bm{t}_k]$, we update the point mapping of each point in ${\bm P}^{(k)}$ to its closest point in $\bm Q$:
\begin{equation}\label{m_update}
{m^{(k)}_i} =   \mathop{\arg\min}_j  \| \bm{R}_k  {\bm p}^{(k)}_i  + \bm{t}_k - {{\bm q}_j }  \|_2^2.
\end{equation}
 
\textbf{Updating validity mask.} 
Based on the current rigid transformation estimate and the updated point mapping, we update the validity of each point mapping by measuring the mismatch of bidirectional flows and the distance between the warped point and the corresponding point:
\begin{equation}\label{cm_update}\small
{ C^{(k)}_i}  =  \text{I}( \|  {\bm f}^{(k)}_i + {\bm b}_{ m^{(k)}_i}   \|_2<\beta_1)  \cdot \text{I}( \|  \bm{R}_k  {\bm p}^{(k)}_i  + \bm{t}_k - {\bm q}_{ m^{(k)}_i}   \|_2< \beta_2). 
\end{equation}

\subsubsection{Output}
\noindent\textbf{Generating pseudo labels and their binary validity mask.}\quad 
After several alternating iterations, we obtain the final rigid transformation estimate for each supervoxel as the optimal rigid transformation.
Following Eq.~(\ref{region_rigid}), we generate pseudo rigid scene flow labels ${\bm {\widehat D}}^{(k)} $ for each supervoxel ${\bm P}^{(k)}$ from the optimal rigid transformation, thereby obtaining the pseudo scene flow labels ${\bm {\widehat D}}$ for the entire point cloud  ${\bm P}$.
Reliable pseudo scene flow labels represent point-wise correspondences between two point clouds. 
Therefore, we also use the validity mask ${\bm C}$ of point mappings derived from the optimal rigid transformations to indicate the validity of pseudo scene flow labels.
The method of our pseudo label generation is sketched in Algorithm~\ref{alg1}.

\subsection{Self-supervised training with pseudo labels and binary validity mask}\label{Loss_func}
Using the generated pseudo labels and their validity mask, we can achieve the self-supervised training of scene flow estimation networks with supervised loss functions.
In this paper, we apply our self-supervised learning method to BiFlow~\cite{cheng2022bi} with a multi-level $l_2$-norm loss. 
Taking BiFlow as an example, one level of the multi-level self-supervised loss function derived by our method can be written as
\begin{equation}\label{loss}
L = \frac{\sum_{i=1}^{N} C_i \|  {\bm f}_i  - {\bm {\widehat d}}_i  \|_2}{\sum_{i=1}^{N} C_i} ,
\end{equation}
where $\bm f_i$ is the predicted scene flow for point $i$, $ {\bm {\widehat d}}_i $ is our generated pseudo label and $C_i$ is the  validity for this pseudo label.
By using the binary validity mask $\bm C$, we mask out invalid pseudo labels, so that the self-supervised learning  will only be guided by reliable pseudo labels.

\begin{table*}
	\footnotesize
	\caption{ Quantitative results on FT3D$_{\mathbf s}$ test set and KITTI$_{\mathbf s}$.
		Compared with self-supervised approaches, our approach achieves state-of-the-art performance on all metrics.
		Especially, our approach is the only self-supervised one that achieves an EPE metric below 5$cm$.
		Without any ground truth scene flow for supervision, our approach even outperforms some supervised ones.}
	\vspace{-3mm}
	\label{table_SOTA1}
	\renewcommand\arraystretch{1.00}	
	\centering
	\resizebox{2.0\columnwidth}{!}{		
		\begin{tabular}{@{\hskip 0.2cm}l@{\hskip 0.1cm}l@{\hskip 0.2cm}|c@{\hskip 0.7cm}c@{\hskip 0.7cm}c@{\hskip 0.7cm}c|c@{\hskip 0.7cm}c@{\hskip 0.7cm}c@{\hskip 0.7cm}c}
			\Xhline{1.2pt}
			& {\multirow{2}{*}{\textbf{Method}} } &  \multicolumn{4}{c|}{\textbf{FT3D$_{\mathbf s}$}} &  \multicolumn{4}{c}{\textbf{KITTI$_{\mathbf s}$}} \\ 
			&		& {\textbf{EPE}~$\downarrow$} & {\textbf{AS}~$\uparrow$} & {\textbf{AR}~$\uparrow$} & { \textbf{Out}~$\downarrow$} & {\textbf{EPE}~$\downarrow$} & {\textbf{AS}~$\uparrow$} & {\textbf{AR}~$\uparrow$} & { \textbf{Out}~$\downarrow$}\\
			\Xhline{1.2pt}
			\multirow{8}{*}{\rotatebox[origin=c]{90}{Supervised}}
			&FlowNet3D~\cite{liu2019flownet3d} &  0.0864&   47.89&  83.99 &  54.64  & 0.1064& 50.65  & 80.11  &  40.03 \\
			&HPLFlowNet~\cite{gu2019hplflownet} &   0.0804 & 61.44  & 85.55 & 42.87  &  0.1169& 47.83 & 77.76& 41.03 \\
			&PointPWC-Net~\cite{wu2020pointpwc}  &  0.0588 & 73.79 & 92.76 &34.24  &  0.0694 & 72.81 & 88.84 &26.48   \\
			&FLOT~\cite{puy2020flot}  &   0.0520&  73.20&  92.70& 35.70 &   0.0560&  75.50&  90.80& 24.20\\
			&FlowStep3D~\cite{kittenplon2021flowstep3d}  & 0.0455  & 81.62 &  96.14& 21.65 & 0.0546  & 80.51 &  92.54& 14.92 \\
			&RCP~\cite{gu2022rcp} &0.0403 &85.67 &96.35 &19.76& 0.0481 &   84.91 & 94.48  & 12.28\\
			&3DFlow~\cite{wang2022matters} & 0.0281  & 92.90 &  98.17 & 14.58 & 0.0309  & 90.47 &  95.80 & 16.12 \\
			&BiFlowNet~\cite{cheng2022bi}  & 0.0280  & 91.80 &  97.80 & 14.30 & 0.0300  & 92.00 &  96.00 & 14.10  \\
			\cline{1-10}
			\multirow{9}{*}{\rotatebox[origin=c]{90}{Self-supervised}}
			&Ego-motion~\cite{tishchenko2020self} &  0.1696 &  25.32  &  55.01 &  80.46 &   0.4154  &  22.09  &  37.21 &  80.96 \\
			&PointPWC-Net~\cite{wu2020pointpwc}  & 0.1213 & 32.39  &  67.42 & 68.78  &  0.2549 & 23.79 & 49.57 & 68.63 \\
			&SLIM$^\dag$~(input 8,192 points)~\cite{baur2021slim}&  $-$ &   $-$ &   $-$ &   $-$  & 0.1207 & 51.78 & 79.56 & 40.24\\
			&Self-Point-Flow~\cite{li2021self}  & 0.1009 & 42.31  &  77.47 & 60.58  & 0.1120 & 52.76  &  79.36 & 40.86  \\
			&FlowStep3D~\cite{kittenplon2021flowstep3d}  & 0.0852 & 53.63 & 82.62 &  41.98  & 0.1021 &   70.80 & 83.94 &  24.53\\
			&PDF-Flow~\cite{he2022self} &0.0750 & 58.90 & 86.20 &47.00 & 0.0920 &74.70 & 87.00 & 28.30 \\
			&RCP~\cite{gu2022rcp}  & 0.0765 & 58.58 & 86.02 & 41.42 & 0.0763 & 78.56 & 89.21 & 18.49\\
			&SLIM$^\dag$~(input all points)~\cite{baur2021slim}&   $-$ &  $-$ &  $-$ &  $-$  & 0.0668 & 76.95 & 93.42 & 24.88\\
			&RigidFlow (using FLOT as model)~\cite{li2022rigidflow}    & 0.0692 	&  59.62 	&  87.10 	&  46.42  & 0.0619 	&  72.37 	&  89.23 &	26.18 \\
			&SPFlowNet~\cite{shen2023self}  &  0.0606 &   68.34 & 90.74 &  38.76  & \bf 0.0362 & 87.24 & \bf 95.79 &17.71\\
			&\textbf{RigidFlow++}  & \bf 0.0477 & \bf 82.83 & \bf95.04 & \bf 27.86 & \bf 0.0363 & \bf 91.73  &\bf  95.51  &\bf 16.04  \\
			\Xhline{1.2pt}
		\end{tabular}
	}
\end{table*}

\begin{table*}
	\footnotesize
	\caption{ Quantitative results on FT3D$_{\mathbf o}$ test set and KITTI$_{\mathbf o}$.
		The scores of BiFlowNet~\cite{cheng2022bi} are obtained using the official implementation$^\dag$.
		Without using ground truth data for training, our method is superior to previous self-supervised methods and performs better than some supervised counterparts.}
	\vspace{-3mm}
	\label{table_SOTA_Occ}
	\renewcommand\arraystretch{1.05}	
	\centering
	\resizebox{2.0\columnwidth}{!}{		
		\begin{tabular}{@{\hskip 0.2cm}l@{\hskip 0.1cm}l@{\hskip 0.0cm}|c|c@{\hskip 0.3cm}c@{\hskip 0.5cm}c@{\hskip 0.5cm}c@{\hskip 0.5cm}c|c@{\hskip 0.3cm}c@{\hskip 0.5cm}c@{\hskip 0.5cm}c@{\hskip 0.2cm}}
			\Xhline{1.2pt}
			& {\multirow{2}{*}{\textbf{Method}}}  & \multirow{2}{*}{\textbf{Training data}} &  \multicolumn{5}{c|}{\textbf{FT3D$_{\mathbf o}$}}&  \multicolumn{4}{c}{\textbf{KITTI$_{\mathbf o}$}}\\ 
			&  & &{\textbf{EPE$_{\rm full}$}~$\downarrow$} & {\textbf{EPE}~$\downarrow$} & {\textbf{AS}~$\uparrow$} & {\textbf{AR}~$\uparrow$} & { \textbf{Out}~$\downarrow$} &{\textbf{EPE$_{\rm full}$}~$\downarrow$} & {\textbf{AS}~$\uparrow$} & {\textbf{AR}~$\uparrow$} & { \textbf{Out}~$\downarrow$}\\
			\Xhline{1.2pt}
			\multirow{4}{*}{\rotatebox[origin=c]{90}{Supervised}}
			&FLOT~\cite{puy2020flot} &   {FT3D$_{\mathbf o}$ training set} & 0.250 & 0.153 & 39.6 & 66.0  & 66.2  & 0.130  & 27.8  &  66.7  & 52.9 \\
			&OGSFNet~\cite{Ouyang_2021_CVPR} & {FT3D$_{\mathbf o}$ training set}  & 0.163 & 0.121  & 55.1 &  77.6  & 51.8 & 0.075 &70.6 &  86.9  & 32.7\\
			&3DFlow~\cite{wang2022matters} & {FT3D$_{\mathbf o}$ training set}  &0.117 & 0.063  & 79.1 &  90.9 & 27.9 &  0.070  &  78.3 &  89.8  & 27.7 \\
			&BiFlowNet$^\dag$~\cite{cheng2022bi}  & {FT3D$_{\mathbf o}$ training set}  & 0.125 & 0.079  & 74.1 &  87.3 & 31.8 &  0.045  &   88.1 &  95.6 & 20.2 \\
			\cline{1-12}
			\multirow{6}{*}{\rotatebox[origin=c]{90}{Self-supervised}}
			&3D-OGFlow~\cite{ouyang2021occlusion}  & {FT3D$_{\mathbf o}$ training set}  & 0.337 &0.279 & 12.3 & 35.9 & 91.0& 0.209   & 21.1 &  49.0  &  72.4 \\
			&\textbf{RigidFlow++}  & {FT3D$_{\mathbf o}$ training set}  & \bf 0.209 & \bf 0.144 &\bf 51.3& \bf 73.6 	&  \bf 56.7 & \bf 0.059 &\bf  82.6 & \bf 92.5 & \bf 23.6 \\
			\cline{2-12}
			&Self-Point-Flow~\cite{li2021self}&  {KITTI$_{\mathbf r}$}  & $-$&  $-$&  $-$&  $-$&  $-$ &   0.115 &   36.7 & 67.1 &   54.3 \\
			&RigidFlow (using FLOT as model)~\cite{li2022rigidflow}&  {KITTI$_{\mathbf r}$}  & $-$&  $-$&  $-$&  $-$&  $-$ &   0.100 &   49.3 & 75.6 &   44.3 \\
			&SPFlowNet~\cite{shen2023self}  &  {KITTI$_{\mathbf r}$}  & $-$&  $-$&  $-$&  $-$&  $-$ &   0.088 & 59.5 &	81.1 &  39.6 \\
			&\textbf{RigidFlow++}&  {KITTI$_{\mathbf r}$}  & $-$&  $-$&  $-$&  $-$&  $-$ & \bf 0.052 &  \bf 87.0 &\bf 93.3	&  \bf 20.1   \\
			\Xhline{1.2pt}
		\end{tabular}
	}
\end{table*}

\begin{figure*}[tb]
	\centering
	\includegraphics[height=9.0cm]{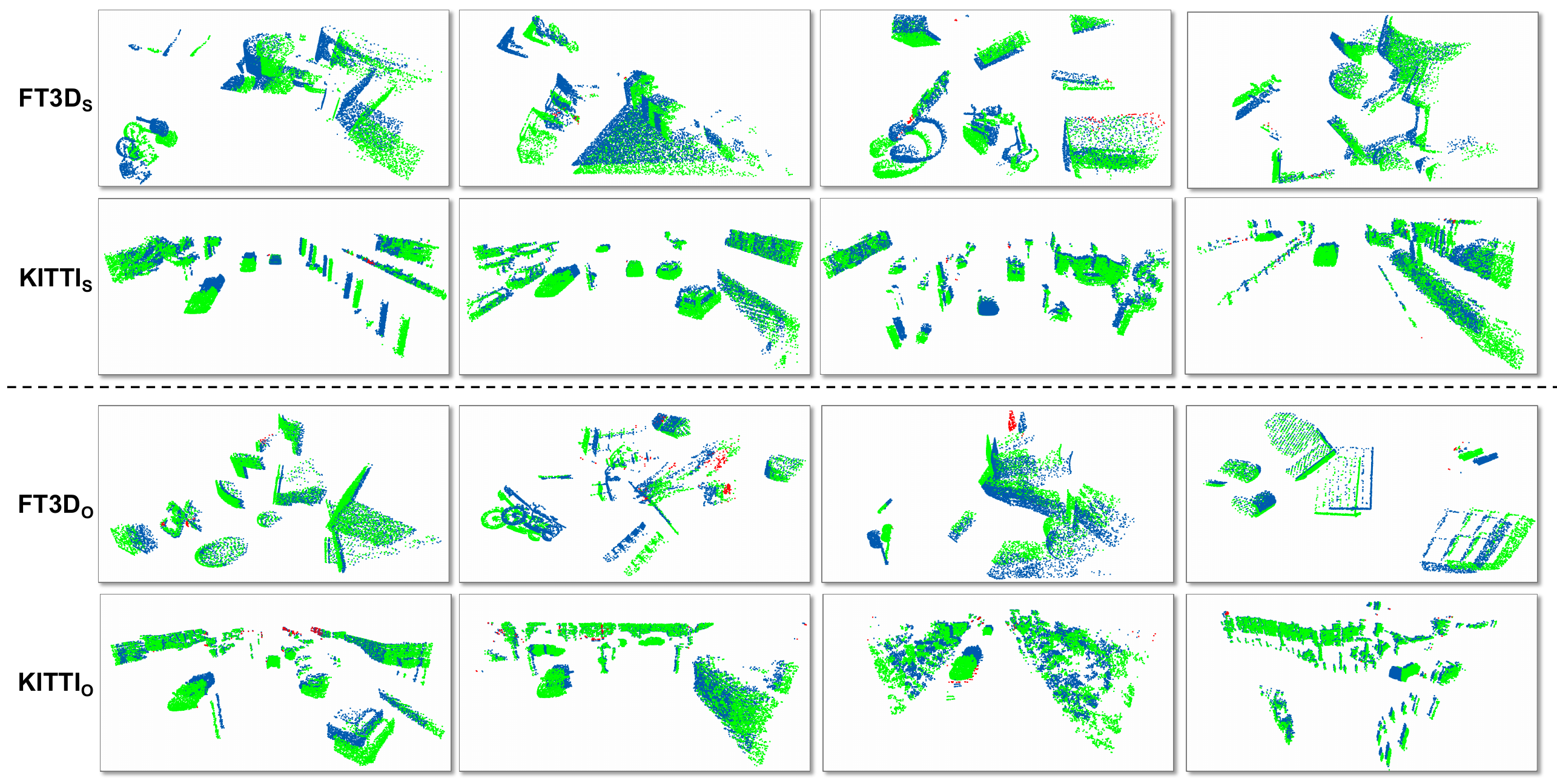}
	\vspace{-3mm}
	\caption{Qualitative results on non-occluded data, \textbf{FT3D$_{\mathbf s}$}  and \textbf{KITTI$_{\mathbf s}$}, and  occluded data, \textbf{FT3D$_{\mathbf o}$}  and \textbf{KITTI$_{\mathbf o}$}. {\color{NavyBlue}\textbf{Blue points}} represent the {\color{NavyBlue}\textbf{source}} point cloud.  {\color{Green}\textbf{Green points}}  represent the points translated by the {\color{Green}\textbf{correct}} scene flow predictions. {\color{Red}\textbf{Red points}} represent the points translated by the {\color{Red}\textbf{incorrect}} predictions. The scene flow predictions are measured by \textbf{AR}. 
	}
	\label{fig_vis}	
\end{figure*}

\section{Experiment}
To validate the effectiveness of our self-supervised learning method,  we first compare our method with the state-of-the-art fully-supervised and self-supervised methods in Sec.~\ref{Sec.SOTA_method}. 
Then, we compare our method with advanced test-time optimization-based methods in Sec.~\ref{Sec.SOTA_TTO}.
In Sec.~\ref{Sec.AS}, we conduct various ablation experiments to analyze the contribution of different components in our method. 
And in Sec.~\ref{Sec.APL}, we design some quantitative and qualitative experiments to evaluate the generated pseudo labels for further analysis.
Finally, we extend our method to the task of self-supervised class-agnostic motion prediction in Sec.~\ref{Sec.Motion}.
All experiments are performed on a large-scale synthetic  FlyingThings3D dataset~\cite{mayer2016large}, a real-world KITTI 2015 dataset~\cite{menze2015object,menze2015joint}, and a large-scale autonomous driving dataset, nuScenes dataset~\cite{caesar2020nuscenes}.
In the following, we introduce the datasets, implementation details, and evaluation metrics in our scene flow estimation experiments.
The details of our motion prediction experiments are contained in Sec.~\ref{Sec.Motion}.

\textbf{Datasets.}\quad 
We conduct scene flow estimation experiments on FlyingThings3D~\cite{mayer2016large} and KITTI 2015~\cite{menze2015object,menze2015joint}. 
3D data are not directly provided by the two original datasets, thus the point clouds need to be extracted from the original data.
Following FLOT~\cite{puy2020flot}, we denote the two point cloud datasets prepared by HPLFlowNet~\cite{gu2019hplflownet} as \textbf{FT3D$_{\mathbf s}$} and \textbf{KITTI$_{\mathbf s}$}, respectively.
For FT3D$_{\mathbf s}$ and KITTI$_{\mathbf s}$, there are no occluded points in the processed point clouds.
We denote the two datasets prepared by FlowNet3D~\cite{liu2019flownet3d} as \textbf{FT3D$_{\mathbf o}$} and \textbf{KITTI$_{\mathbf o}$}, respectively, where occluded points are preserved.
Specially, following~\cite{puy2020flot},  we removed 7 training samples in which all points were occluded from the training set of FT3D$_{\mathbf o}$.
FlowNet3D~\cite{liu2019flownet3d} also splits the KITTI$_{\mathbf o}$ data to use the first 100 pairs for finetuning and the rest 50 pairs for testing.
Here, we denote the finetuning part as \textbf{KITTI$_{\mathbf f}$} and the rest testing data as \textbf{KITTI$_{\mathbf t}$}.
Following the raw data sampling strategy used in~\cite{li2021self}, we extract some raw point clouds from KITTI dataset as training samples (6,026 pairs) and denote them as \textbf{KITTI$_{\mathbf r}$}. There is no overlap between KITTI$_{\mathbf r}$ and KITTI$_{\mathbf o}$.

\textbf{Implementation details.}\quad 
During the evaluation of our method for scene flow estimation, we perform experiments on two types of data: point clouds without occlusions and point clouds with occlusions.
For the experiment on point clouds without occlusions, we follow the experimental setting in~\cite{gu2019hplflownet, wu2020pointpwc, kittenplon2021flowstep3d}.
Specifically, we train a BiFlowNet~\cite{cheng2022bi} model by our self-supervised approach on FT3D$_{\mathbf s}$ training set (19,640 pairs) and test it on FT3D$_{\mathbf s}$ test set (3,824 pairs) and KITTI$_{\mathbf s}$ (142 pairs).
For a pair of point clouds, we randomly sample 8,192 points in each point cloud as input.
In the pseudo label generation phase, we decompose the source point cloud into 30 supervoxels with an over-segmentation method~\cite{lin2018toward} and set the iteration number in our piecewise pseudo label generation module to 4.
And we set the threshold values $\beta_1$ and $\beta_2$ to  $0.2m$  and  $0.1m$, respectively, and the kernel’s bandwidth parameter $\theta^2$ to $0.005$. 
Specifically, we start confidence reweighting and validity reasoning after 10 epochs.
In the first 10 epochs, we fix the confidence weight and the validity mask  to $1$.
We set the batchsize to 8 and use Adam optimizer~\cite{kingma2014adam} with an initial learning rate of 0.001.

For the experiment on point clouds with occlusions, we first train a BiFlowNet model on synthetic FT3D$_{\mathbf o}$ training set (19,999~pairs) using our self-supervised method and evaluate it on FT3D$_{\mathbf o}$ test set (2,003 pairs), KITTI$_{\mathbf o}$ (150 pairs), and KITTI$_{\mathbf t}$ (50 pairs).
Then, we train another BiFlowNet model on realistic KITTI$_{\mathbf r}$ and evaluate it on KITTI$_{\mathbf o}$.
The settings for the input and the pseudo label generation in this experiment are the same as those used in the experiment on point clouds without occlusions.

\textbf{Evaluation metrics.}\quad
When evaluating our method  on scene flow estimation,
we adopt five evaluation metrics used in \cite{cheng2022bi}.
We denote the ground truth scene flow and predicted scene flow as $\bm D$ and $\bm F$, respectively.
The metrics are defined as follows:
\textbf{EPE$_{\rm full}$}(m): $\| \bm D - \bm F\|_2$, end point error, averaged over all points;
\textbf{EPE}(m): end point error, averaged over non-occluded points;
\textbf{AS}(\%): the ratio of points with EPE $<$ 0.05m or relative error $< 5\%$;
\textbf{AR}(\%): the ratio of points with EPE $<$ 0.1m or relative error $< 10\%$;
\textbf{Out}(\%): the ratio of points with EPE $>$ 0.3m or relative error $> 10\%$.

\begin{table*}
		\vspace{-1mm}
	\caption{Comparison with test-time optimization-based methods. Experiments are conducted with 8,192 points as input. Our self-supervised learning method  achieves a good trade-off between performance and inference time.}
	\vspace{-3mm}
	\label{TTO}
	\centering
	\renewcommand\arraystretch{1.0}
	\resizebox{2.0\columnwidth}{!}{	
	\begin{tabular}{l@{\hskip 0.8cm}|@{\hskip 0.1cm}c@{\hskip 0.1cm}|c@{\hskip 0.2cm}@{\hskip 0.2cm}c@{\hskip 0.2cm}@{\hskip 0.2cm}c|c@{\hskip 0.2cm}@{\hskip 0.2cm}c@{\hskip 0.2cm}@{\hskip 0.2cm}c|c@{\hskip 0.1cm}}
		\Xhline{1.6pt}
		\multirow{2}{*}{\textbf{Method}} &\multirow{2}{*}{\textbf{Type}} &     \multicolumn{3}{c|}{\textbf{FT3D$_{\mathbf o}$}} & \multicolumn{3}{c|}{\textbf{KITTI$_{\mathbf t}$}} & \multirow{2}{*}{\textbf{Inference time (s)}} \\ 
		
		&  &  {  \textbf{EPE}~$\downarrow$}& {\textbf{AS}~$\uparrow$}&  {\textbf{AR}~$\uparrow$}&
		{\textbf{EPE}~$\downarrow$} & {\textbf{AS}~$\uparrow$} & {\textbf{AR}~$\uparrow$} & \\
		\Xhline{1.6pt}
		NSFP~\cite{li2021neural} & Optimization-based &	  0.183&   40.4  &  66.5 & 0.035 & 93.1 & \bf 96.9 & $\sim 6$\\
		SCOOP~\cite{lang2022scoop}   &  Self-supervised learning + Optimization-based & 0.275 & 24.4& 49.1   & \bf 0.028 & \bf 95.2 & \bf 96.9 & $\sim 3$\\
		\textbf{RigidFlow++} &     Self-supervised learning-based&  \bf 0.144 & \bf 51.3   & \bf 73.6 & 0.063 & 83.8 & 92.0 &\bf{$\sim$ 0.2} \\				
		\Xhline{1.6pt}
	\end{tabular}
}
\end{table*}

\subsection{Comparison with State-of-the-art Learning-based Scene Flow Methods}\label{Sec.SOTA_method}

\subsubsection{Results on FT3D$_{ \rm s}$ and KITTI$_{ \rm s}$} 
We evaluate our self-supervised learning method on non-occluded FT3D$_{\mathbf s}$ test set and KITTI$_{\mathbf s}$ data, following the experimental setting in~\cite{gu2019hplflownet, wu2020pointpwc, kittenplon2021flowstep3d}.
The results, presented in Table~\ref{table_SOTA1}, reveal that our method outperforms the competing self-supervised approaches across all metrics. 
This demonstrates the effectiveness and generalization ability of our self-supervised learning algorithm. 
Notably, our method is the only self-supervised approach that achieves an \textbf{EPE} metric below 5$cm$ on the two datasets.
For this metric, our method outperforms the original version, RigidFlow~\cite{li2022rigidflow}, and the recent SPFlowNet~\cite{shen2023self} by 31\% and 21\% on FT3D$_{ \mathbf s}$, respectively.

We also compare our self-supervised method with advanced supervised approaches that are trained on FT3D$_{ \mathbf s}$ training set.
As shown in Table~\ref{table_SOTA1}, without any ground truth for supervision, our self-supervised method performs better than supervised  FlowNet3D~\cite{liu2019flownet3d},  HPLFlowNet~\cite{gu2019hplflownet},  PointPWC-Net~\cite{wu2020pointpwc}, and FLOT~\cite{puy2020flot} on FT3D$_{ \mathbf s}$.
Evaluated on KITTI$_{ \mathbf s}$ without fine-tuning, our self-supervised method also achieves better generalization ability than the above four supervised approaches. 
Some qualitative results on FT3D$_{ \mathbf s}$ and KITTI$_{ \mathbf s}$ are shown in Fig.~\ref{fig_vis}.

\subsubsection{Results on FT3D$_{ \rm o}$ and KITTI$_{ \rm o}$ }\label{Occ_experiment}
We then evaluate our self-supervised learning method on occluded FT3D$_{ \rm o}$ test set and KITTI$_{\rm o}$ data.
Specifically, in KITTI$_{\rm o}$ data, following previous works~\cite{liu2019flownet3d,Ouyang_2021_CVPR,cheng2022bi,ouyang2021occlusion}, we remove the ground points for evaluation.
The reported scores on KITTI$_{\rm o}$ for~\cite{li2021self,li2022rigidflow,shen2023self,puy2020flot,wang2022matters} in Table~\ref{table_SOTA_Occ}  are different from the scores in their own papers; this is because these papers only evaluate on points with depth less than 35m.  

We first train a BiFlowNet model with our self-supervised  method on FT3D$_{ \rm o}$ training set.
As presented in Table~\ref{table_SOTA_Occ}, our method outperforms self-supervised 3D-OGFlow~\cite{ouyang2021occlusion} by a large margin on all metrics.
Specifically, on the  \textbf{EPE} metric, our method  achieves a 48\% error reduction  on FT3D$_{ \rm o}$ test set and a 71\% error reduction on KITTI$_{\rm o}$ data when compared to 3D-OGFlow.
Moreover, without using any ground truth data for training, our self-supervised method outperforms supervised FLOT~\cite{puy2020flot} on FT3D$_{ \rm o}$ test set and exhibits better generalization ability than supervised OGSFNet~\cite{Ouyang_2021_CVPR}, 3DFlow~\cite{wang2022matters}, and FLOT  on  KITTI$_{\rm o}$.
Qualitative results on FT3D$_{ \mathbf o}$ and KITTI$_{ \mathbf o}$ are shown in Fig.~\ref{fig_vis}.

Then, we train another BiFlowNet model with our self-supervised method on unlabeled KITTI$_{ \rm r}$ data and evaluate on KITTI$_{ \rm o}$. 
As shown in Table~\ref{table_SOTA_Occ}, compared with self-supervised Self-Point-Flow~\cite{li2021self}, RigidFlow~\cite{li2022rigidflow}, and SPFlowNet~\cite{shen2023self}, our method achieves the best performance on all metrics.
Specifically, our method outperforms the original version, RigidFlow, and the recent SPFlowNet by 48\% and 40\% on the \textbf{EPE} metric, respectively.
It is worth noting that the BiFlowNet trained on unlabeled KITTI$_{\rm r}$ via our self-supervised method performs on par with the fully supervised BiFlowNet~\cite{cheng2022bi} trained on labeled FT3D$_{\rm o}$ training set, despite FT3D$_{\rm o}$ contains more training samples than KITTI$_{\rm r}$ (20K  $v.s.$ 6K). 
This demonstrates the advantage of our self-supervised learning strategy, i.e., the proposed self-supervised learning allows models to learn useful representations directly from unannotated real-world data.

\begin{table*}
	\footnotesize
	\caption{ Ablation study for the generalization ability of our self-supervised training strategy to 3DFlow~\cite{wang2022matters} on  non-occluded and occluded FT3D and KITTI.}
	\vspace{-3mm}
	\label{table_Generalization}
	\renewcommand\arraystretch{1.0}	
	\centering
	\resizebox{2.0\columnwidth}{!}{		
		\begin{tabular}{l|cc|cc||ccc|cc}
			\Xhline{1.2pt}
			{\multirow{2}{*}{\textbf{Network Backbone}}}  &  \multicolumn{2}{c|}{\textbf{FT3D$_{\mathbf s}$}}&  \multicolumn{2}{c||}{\textbf{KITTI$_{\mathbf s}$}} &  \multicolumn{3}{c|}{\textbf{FT3D$_{\mathbf o}$}}&  \multicolumn{2}{c}{\textbf{KITTI$_{\mathbf o}$}}  \\ 
			&{\textbf{EPE}~$\downarrow$} &{\textbf{AS}~$\uparrow$}&{\textbf{EPE}~$\downarrow$}    &{\textbf{AS}~$\uparrow$}  &{\textbf{EPE$_{\rm full}$}~$\downarrow$} &{\textbf{EPE}~$\downarrow$} &{\textbf{AS}~$\uparrow$}  
			&{\textbf{EPE$_{\rm full}$}~$\downarrow$} &{\textbf{AS}~$\uparrow$} \\
			\hhline{-----||-----}
			3DFlow~\cite{wang2022matters} (Self-supervised training by RigidFlow++)&  0.062& 73.3 & 0.085 &  72.6  & 0.248 & 0.146 & 44.5 & 0.088 & 69.8 \\
			\Xhline{1.2pt}
		\end{tabular} 
	}
\end{table*}

\begin{table*}
	\caption{Ablation study for confidence-aware piecewise pseudo label generation module. $\rm \Delta$ denotes the difference in metric with respect to the baseline.}
	\vspace{-3mm}
	\label{Table_ablation}
	\centering
	\renewcommand\arraystretch{1.05}
	\resizebox{2.0\columnwidth}{!}{			
		\begin{tabular}{l@{\hskip 0.2cm}|c|c@{\hskip 0.4cm}c|c@{\hskip 0.4cm}c|c@{\hskip 0.5cm}c||c@{\hskip 0.5cm}c@{\hskip 0.05cm}}
			\Xhline{1.6pt}
			\multirow{2}{*}{Method} & \multirow{2}{*}{NN search} &  {Region-wise  }&  Region-wise & Confidence score& Validity&     \multicolumn{2}{c||}{\textbf{FT3D$_{\mathbf s}$}} & \multicolumn{2}{c}{\textbf{FT3D$_{\mathbf o}$}}  \\
			
			&  &  center alignment & rigid alignment &   from consistency& mask reasoning&   {  \textbf{EPE}~$\downarrow$}& { $\rm \Delta$\textbf{EPE}}&
			{\textbf{EPE}~$\downarrow$} & {$\rm \Delta$\textbf{EPE}} \\
			\hhline{--------||--}
			A (\bf Baseline) &   \checkmark & &  & & & 0.242 &  \green{0.000}   & 0.734  & \green0.000 \\
			\hhline{--------||--}
			B &   \checkmark & \checkmark&  & & & 0.100&  \green-~0.142    & 0.225  & \green-~0.509  \\
			C (\bf RigidFlow) &   \checkmark & &  \checkmark& & &	0.059 & \green-~0.183   & 0.210  &  \green-~0.524  \\
			D &   \checkmark & &  \checkmark& \checkmark& &		0.058 &  \green-~0.184   &  0.173  &  \green-~0.561 \\
			E &   \checkmark & &  \checkmark& &\checkmark &	0.051 &  \green-~0.191  & 0.167 & \green-~0.567  \\				
			F (\bf RigidFlow++) &   \checkmark & &  \checkmark&\checkmark &\checkmark &	\bf 0.047 &  \bf \green-~0.195   &  \bf 0.144  & \bf \green -~0.590\\
			\Xhline{1.6pt}
		\end{tabular}
	}
\end{table*}

\subsection{Comparison with Test-time Optimization-based Scene Flow Methods}\label{Sec.SOTA_TTO}
When ground truth data is unavailable, in addition to training scene flow estimation models via self-supervised learning, estimating scene flow at test time by optimization is an alternative. 
To validate the effectiveness of our self-supervised learning method, we compare it with two advanced test-time optimization-based methods: NSFP~\cite{li2021neural} and SCOOP~\cite{lang2022scoop}.
Specifically,  NSFP is a pure optimization-based method, where scene flow is optimized for each scene separately with a neural network as an implicit regularizer.
And SCOOP combines self-supervised learning and optimization.
In SCOOP, a self-supervised correspondence model is used to estimate initial scene flow, and a test-time optimization module is used to optimize residual flow refinement.

Following NSFP and SCOOP, we test the three methods on  FT3D$_{ \rm o}$ test set and KITTI$_{ \rm t}$. 
For our self-supervised method, we train a BiFlowNet model with it on FT3D$_{ \rm o}$ training set and directly evaluate the model on the two test data sets.
For NSFP and SCOOP, we follow their experimental protocols in their papers, but set the number of input points to 8,192.
Therefore, the scores differ from those reported in their own papers, as these papers evaluate the methods with 2,048 points as input.
As presented in Table~\ref{TTO}, although NSFP and SCOOP perform better than our method on KITTI$_{ \rm t}$, ours surpasses the two methods by a large margin on FT3D$_{ \rm o}$ test set and has significantly faster inference (15$\times$ faster).
Therefore, compared to the two test-time optimization-based methods, our self-supervised learning method achieves a good trade-off between performance and inference time, which makes it more suitable for time-sensitive and low-power applications.

\subsection{Ablation study}\label{Sec.AS}
We first evaluate the generalization ability of our method across scene flow models.
Then, we conduct experiments to validate the effectiveness of each component in our method.
Afterward, we analyze the impact of the number of supervoxels and the number of update iterations on our method.
Finally, we discuss the time consumption of our method during training.
In this section, unless otherwise specified, we adopt the BiFlowNet~\cite{cheng2022bi} as our scene flow estimation model.

\textbf{Generalization ability across different scene flow models.}\quad 
Our self-supervised training method is efficient for different scene flow estimation models.
Apart from applying our method to BiFlowNet~\cite{cheng2022bi} (as presented in Table~\ref{table_SOTA1} and Table~\ref{table_SOTA_Occ}), we also apply it to another advanced scene flow estimation model, 3DFlow~\cite{wang2022matters}, to analyze the generalization ability of our method. 
Following the same evaluation strategy in Sec.~\ref{Sec.SOTA_method}, we employ our RigidFlow++ to train one 3DFlow model on FT3D$_{\mathbf s}$ training set and FT3D$_{\mathbf o}$ training set, respectively.
As shown in Table~\ref{table_Generalization}, for the experiment on non-occluded data, the 3DFlow trained on FT3D$_{\mathbf s}$ achieves  an \textbf{EPE} of $0.062m$ on FT3D$_{\mathbf s}$ test set and an \textbf{EPE} of $0.085m$ on KITTI$_{\mathbf s}$ without fine-tuning, which outperforms self-supervised PDF-Flow~\cite{he2022self}, FlowStep3D~\cite{kittenplon2021flowstep3d}, and Self-Point-Flow~\cite{li2021self} (as shown in Table~\ref{table_SOTA1}).
For the experiment on occluded data, the 3DFlow trained on FT3D$_{\mathbf o}$  achieves  an \textbf{EPE} of $0.146m$ on FT3D$_{\mathbf o}$ test set and an \textbf{EPE} of $0.088m$ on KITTI$_{\mathbf o}$ without fine-tuning, which performs significantly better than self-supervised 3D-OGFlow~\cite{ouyang2021occlusion} (as shown in Table~\ref{table_SOTA_Occ}). 
The experimental results show the generalization ability of our self-supervised training method to different scene flow models.

\textbf{Confidence-aware piecewise pseudo label generation module.}\quad 
At the core of our framework is the confidence-aware piecewise pseudo label generation module. 
Specifically, this module generates pseudo labels and their validity mask by explicitly enforcing region-wise rigid alignments and employing a confidence mechanism to suppress outliers.
In the following, we analyze the advantages of the two designs separately and conduct experiments on both occluded FT3D$_{\mathbf o}$ and non-occluded FT3D$_{\mathbf s}$.

Firstly, to analyze the impact of region-wise  rigid alignments on pseudo scene flow generation, we design three competing methods: 
\begin{itemize}
	\item[\textbf{A}]  
	Nearest point alignment, i.e., nearest neighbor search. For each point, we directly treat the initial match derived from our point mapping initialization (Eq.~\ref{m_initial})  as the corresponding point to produce a pseudo label without considering any region-wise constraints.
	
	\item[\textbf{B}]  
	Region-wise  center alignment. When updating transformation,  we only encourage the center of each supervoxel to coincide with that of its counterpart rather than enforcing region-wise  rigid alignments. Therefore,  we fix the rotation matrix $\bm{R}_k$ to an identity matrix.
	
	\item[\textbf{C}]    
	Region-wise rigid alignment. This method  is our original version, RigidFlow, which encourages region-wise  rigid alignments but treats all point mappings and pseudo labels as confident and valid.
\end{itemize}
Note that, we do not include our confidence mechanism to the three methods.
Therefore, we fix the confidence weight $w^{(k)}_i$ to $1$ and the validity mask ${ C^{(k)}_i}$ to $1$ in the methods.
As presented in Table~\ref{Table_ablation}, for the FT3D$_{\mathbf s}$ test set, \textbf{C} outperforms \textbf{A} and \textbf{B} by 75\% and 41\% on the \textbf{EPE} metric, respectively.
And for the FT3D$_{\mathbf o}$ test set, \textbf{C} outperforms \textbf{A} and \textbf{B} by 71\% and 6\% on the \textbf{EPE} metric, respectively.
The results demonstrate the superior performance of region-wise rigid alignment compared to nearest point alignment and region-wise center alignment for pseudo label generation.

Secondly, regarding the method of region-wise rigid alignment~(\textbf{C}) as a basic strategy, we analyze the impact of our designed confidence mechanism. 
In this mechanism, we reason about the validity of point mapping and pseudo labels by the constraints of forward-backward consistency and spatial proximity.
Then, for the valid point mappings, we measure their confidence weights by the forward-backward consistency, and for the invalid point mappings, we set their confidence weights to 0.
Specifically, we design three competing methods: 
\begin{itemize}
	\item[\textbf{D}]    
 	Region-wise rigid alignment with confidence reweighting by consistency only. In this method, we fix the validity mask ${ C^{(k)}_i}$ to 1. Accordingly, the confidence weight $w^{(k)}_i$ is equal to the confidence score  ${\bar w}^{(k)}_i$  in Eq.~(\ref{weights}).
 	
	\item[\textbf{E}]    
	Region-wise rigid alignment with validity reasoning. In this method, the confidence score ${\bar w}^{(k)}_i$ is fixed to 1  in Eq.~(\ref{coarseweights}).  Accordingly, the confidence weight $w^{(k)}_i$ is equal to the validity mask ${ C^{(k)}_i}$  in Eq.~(\ref{weights}).

	\item[\textbf{F}]    
	Region-wise rigid alignment with confidence reweighting and validity reasoning. This method is our RigidFlow++.
 	
\end{itemize}
As presented in Table~\ref{Table_ablation}, compared with RigidFlow, using confidence score from forward-backward consistency to reweight point mappings reduces the \textbf{EPE} from $0.210m$ to $0.173m$ on FT3D$_{\mathbf o}$.
Additionally, by reasoning about the validity for point mappings and pseudo labels, we observe a decrease in the \textbf{EPE} from $0.059m$ to $0.051m$ on FT3D${\mathbf s}$, as well as a reduction from $0.210m$ to $0.167m$ on FT3D${\mathbf o}$.
Finally, by incorporating both validity reasoning and confidence reweighting into region-wise rigid alignment, our RigidFlow++ achieves a further reduction in \textbf{EPE} to $0.047m$ on FT3D${\mathbf s}$ and $0.144m$ on FT3D${\mathbf o}$, which
shows the effectiveness of our confidence mechanism.

Specifically, on the \textbf{EPE} metric, our method outperforms the original version, RigidFlow, by 18\% on FT3D$_{ \rm s}$ and 31\% on FT3D$_{\mathbf o}$, which demonstrates that our RigidFlow++ achieves substantial improvements over the original RigidFlow, especially in occluded scenes.
Table~\ref{table_ablation_iter} shows the impact of the number of update iterations on this module.

\begin{table}
	\caption{ The impact of different update iteration numbers. We report the \textbf{EPE}~$\downarrow$ metric of our method on both FT3D$_{\mathbf s}$ and FT3D$_{\mathbf o}$ test sets.}
	\vspace{-3mm}
	\label{table_ablation_iter}
	\renewcommand\arraystretch{1.05}	
	\centering	
	\begin{tabular}{c|c@{\hskip 0.4cm}c@{\hskip 0.4cm}c@{\hskip 0.4cm}c}
		\Xhline{1.2pt}
		{Iteration number} & {1}  & {2} & {3} & {4}\\
		\Xhline{1.2pt}
		{ FT3D$_{\mathbf s}$} & {0.052 }  & {0.048 } & {0.048 } & {\bf  0.047 }\\ 
		\hline
		{ FT3D$_{\mathbf o}$} & {0.152 }  & {0.144 } & {0.144 } & {\bf  0.144 }\\ 
		\Xhline{1.2pt}
	\end{tabular}
\end{table}

\begin{table}\scriptsize
	\caption{ The impact of different supervoxel numbers. We report the \textbf{EPE}~$\downarrow$ metric of our method on both FT3D$_{\mathbf s}$ and FT3D$_{\mathbf o}$ test sets.}
	\vspace{-3mm}
	\label{table_ablation_sp}
	\renewcommand\arraystretch{1.05}	
	\centering	
	\begin{tabular}{c|c@{\hskip 0.4cm}c@{\hskip 0.4cm}c@{\hskip 0.4cm}c@{\hskip 0.4cm}c}
		\Xhline{1.2pt}
		{Desired supervoxel number} & {10}  & {30} & {50} & {70}  & {90} \\
		\Xhline{1.2pt}
		{FT3D$_{\mathbf S}$} & {0.057 }  & {\bf0.047} & {0.048  } & {0.049 } & {0.052 }\\ 
		\hline
		{FT3D$_{\mathbf O}$} & {\bf 0.143}  & {0.144} & {0.149 } & {0.157}  & {0.154}\\ 
		\Xhline{1.2pt}
	\end{tabular}
\end{table}

\textbf{Impact of supervoxel number.}\quad 
When generating pseudo labels, we decompose a scene into a set of supervoxels and find the rigid motion of each supervoxel.
As shown in Table~\ref{table_ablation_sp}, the model achieves good performance in both occluded and non-occluded scenes when our self-supervised method decomposes each scene into 30 supervoxels for pseudo label generation.

\textbf{Time consumption.}\quad 
We evaluate the running time of our method for a training sample with 8192 points in each point cloud.
When we apply our method to a BiFlowNet model~\cite{cheng2022bi}, the total time consumption of pseudo label generation for a training sample is around 0.5 seconds on a single 2080ti GPU.
Note that our self-supervised method is only performed in the training stage of models and thus does not introduce any extra runtime to the models in the inference stage.

\begin{figure}[t]
	\centering
	\includegraphics[height=3.7cm]{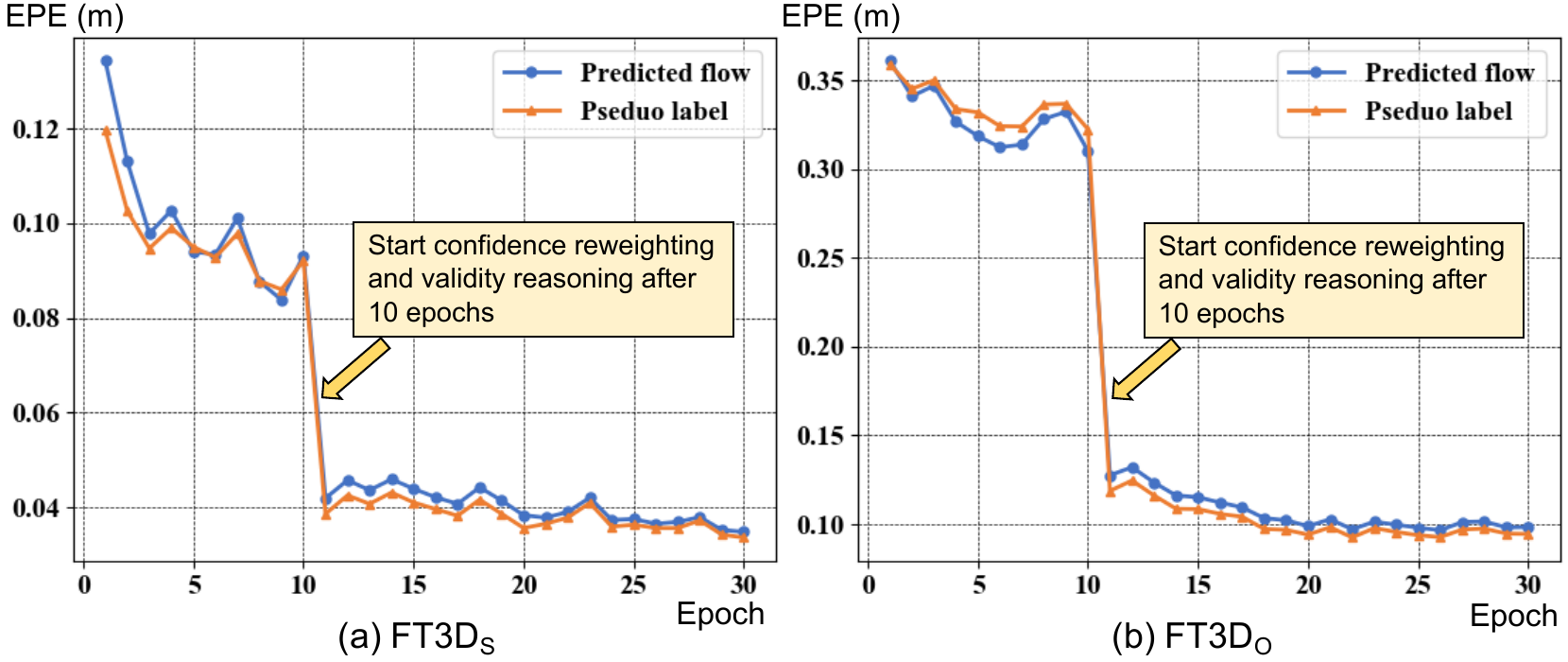}
	\vspace{-2mm}
	\caption{The end point error (\textbf{EPE}, the lower, the better) of model predictions ({\color{NavyBlue}\textbf{Blue curve}}) and generated pseudo labels ({\color{YellowOrange}\textbf{Orange curve}}) on some training samples.  
		During the training, the accuracy of pseudo labels is mostly higher than that of predicted flow. This allows us to apply the pseudo labels as supervision. 
	}
	\label{fig_EPE}	
\end{figure}

\begin{figure*}[htb]
	\centering
	\includegraphics[height=2.55cm]{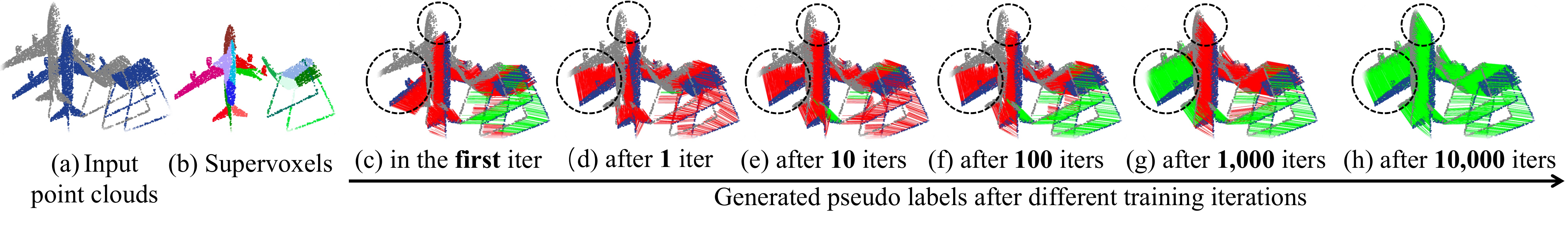}
		\vspace{-2mm}
	\caption{Our generated pseudo labels for the same scene after different training iterations.
		(a) Input point clouds.  {\color{NavyBlue}\textbf{Blue points}} are the {\color{NavyBlue}\textbf{source}} point cloud and {\color{Gray}\textbf{gray points}} are the {\color{Gray}\textbf{target}}.
		(b) Supervoxel results of the source point cloud. Different colors indicate different supervoxels.
		(c) - (h) Our generated pseudo labels after different training iterations.
		{\color{Green}\textbf{Green line}}  indicates the {\color{Green}\textbf{correct}} pseudo label measured by \textbf{AR}. {\color{Red}\textbf{Red line}} indicates the {\color{Red}\textbf{incorrect}} pseudo label.
		The quality of generated pseudo labels for the airplane and the chair is gradually improved along with training iterations.}
	\label{pseudo_label}	
\end{figure*}

\begin{figure*}[htb]
	\centering
	\includegraphics[height=6.1cm]{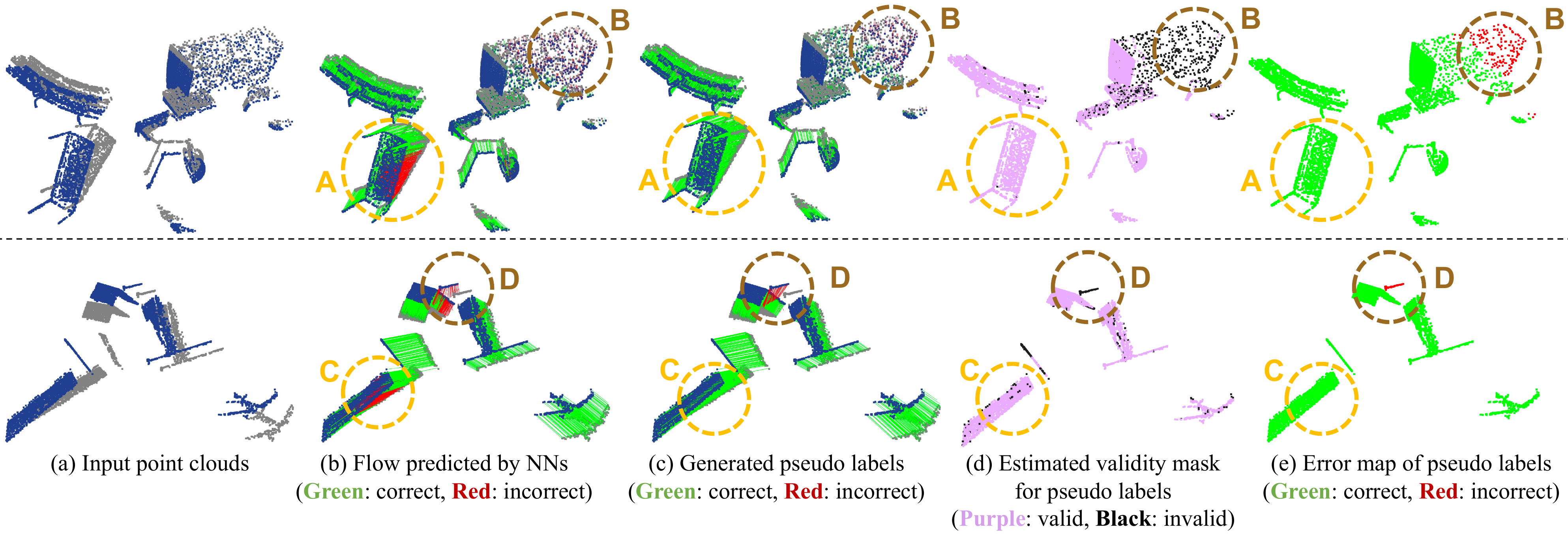}
	\vspace{-3mm}
	\caption{ Visualization of pseudo labels and their validity mask.
		(a) Input point clouds. {\color{NavyBlue}\textbf{Blue points}} are the {\color{NavyBlue}\textbf{source}} point cloud and {\color{Gray}\textbf{gray points}} are the {\color{Gray}\textbf{target}}.
	(b) Predicted scene flow from neural networks being trained.
	(c) Our generated pseudo labels.
	In (b) and (d), green line indicates the correct flow or pseudo label  measured by \textbf{AR}, while red line indicates the incorrect flow or pseudo label.
	(d) Binary validity mask for  pseudo labels estimated by our method. Green point means that its pseudo label is estimated to be valid, while  black point means that its pseudo label is estimated to be invalid.
	(e) Error map of pseudo labels. Green point means that its pseudo label is correct, while red point means that its pseudo label is incorrect. The pseudo labels are measured by \textbf{AR}. 
	In Region A and C, our generated pseudo labels are more accurate than the predicted flow, which allows the pseudo labels to serve as supervision.
	In Region B and D, the inaccurate pseudo labels are estimated to be invalid by our validity reasoning. Based on the validity estimation,  these invalid pseudo labels will be filtered out and the valid ones will dominate the self-supervised training.}
	\label{confidence}	
\end{figure*}

\subsection{Analysis on pseudo labels}\label{Sec.APL}
We conduct some quantitative and qualitative experiments to evaluate the generated pseudo labels and their validity mask for further analysis. 

In Fig.~\ref{fig_EPE},  using part of training samples (197 samples in FT3D$_{ \mathbf s}$, 200 samples in FT3D$_{ \mathbf o}$) as test data, we compare the error of our generated pseudo labels and the scene flow predictions from the neural network being trained.
Specifically, in the first 10 epochs, we evaluate the errors on all points since we do not start validity reasoning and the pseudo labels of all points are used for supervision.
After 10 epochs, we start validity reasoning and only use valid pseudo labels for training, thus, we evaluate the errors on points with valid pseudo labels. 
From Fig.~\ref{fig_EPE}, we can make the following observations. 
(1) The quality of our generated pseudo labels is gradually improved during the training.
(2) Confidence reweighting and validity reasoning lead to a significant improvement in the quality of pseudo labels.
(3) The accuracy of pseudo labels is mostly higher than that of predicted flow. This allows us to apply the pseudo labels as supervision. 
(4) The performance gap between pseudo labels and network predictions is gradually reduced.

Fig.~\ref{pseudo_label} shows the pseudo labels for the same scene after different training iterations.
As illustrated in Fig.~\ref{pseudo_label} (c)-(h), the quality of our generated pseudo labels for the airplane and the chair is gradually improved along with training iterations, which demonstrates the effectiveness of our pseudo label generation method.
Fig.~\ref{confidence} visualizes the pseudo labels and their validity mask for some training samples.
As shown in Fig.~\ref{confidence}(b) and (c), in Region A and  C, our generated pseudo labels are more accurate than the flow predictions, which allows the pseudo labels to serve as supervision.
As shown in Figure ~\ref{confidence}(c), (d), and (e), in Region B and D, although our generated pseudo labels are inaccurate, these pseudo labels are estimated to be invalid by our validity reasoning.
In our method, these invalid pseudo labels will be filtered out and the valid ones will dominate the self-supervised training.

\subsection{Application  on  Self-supervised Class-agnostic Motion Prediction}\label{Sec.Motion}
Given point clouds from past and current frames, motion prediction can be performed by generating a future motion field of all points in the current frame.
By regarding the future motion field as scene flow, we generate pseudo motion labels via our self-supervised scene flow method and use these pseudo labels to train motion prediction models in a self-supervised manner.
In the following, we present the experimental details and the comparisons with state-of-the-art methods.

\textbf{Datasets.}\quad 
We conduct motion prediction experiments on nuScenes~\cite{caesar2020nuscenes}. 
Following previous works~\cite{wu2020motionnet,luo2021self,wang2022sti,li2023weakly}, we  divide the dataset into three parts: 500 scenes for training, 100 for validation, and 250 for test.
During the validation and testing phases, the ground truth motion data is derived from the detection and tracking annotations provided by nuScenes.

\textbf{Implementation details.}\quad 
During the evaluation of our method for motion prediction, we follow ~\cite{wu2020motionnet,wang2022sti} to prepare data.
Specifically, we crop each input point cloud within the spatial bounds of $[-32, 32]\times[-32, 32] \times [-3, 2]$ meters and subsequently partition the input data into voxels with the shape of $(0.25, 0.25, 0.4)$ meters.
And then, we train a motion prediction model by our self-supervised method on nuScenes.
This model is composed of a backbone network and a motion prediction head.
And we utilize the same backbone network as MotionNet~\cite{wu2020motionnet} and employ two-layer 2D convolutions as the motion prediction head.
Following previous works~\cite{luo2021self,li2023weakly}, the input of this model is a point cloud sequence, which consists of 4 point clouds from the past frames and 1 point cloud from the current frame.
And the output is the displacement field for the next 0.5s.
Therefore, when applying our self-supervised method to this model, the source data corresponds to the point cloud in the current frame, while the target data corresponds to the point cloud in the next 0.5s.

When generating pseudo labels, since the ego-motion is compensated to these point clouds, we estimate ground points by RANSAC-based plane fitting and treat these points as static.
Accordingly, we set the pseudo motion labels of ground points to zero, and only apply our pseudo label generation method to the remaining points.  
In pseudo label generation, we decompose the remaining source points into 60 supervoxels and set the iteration number of pseudo label generation module to 2.
And we set the threshold values $\beta_1$ and $\beta_2$ to  $3.0m$  and  $1.0m$, respectively, and the kernel’s bandwidth parameter $\theta^2$ to $0.5$. 
Specifically, we start confidence reweighting and validity reasoning after 20 epochs.
We set the batchsize to 8 and use Adam optimizer with an initial learning rate of 0.0005.

\textbf{Evaluation metrics.}\quad
When evaluating our method  on motion prediction, following~\cite{wu2020motionnet,luo2021self}, we split non-empty cells into three groups: static, slow ($\leq 5{\rm m/s}$), fast ($\geq 5{\rm m/s}$) and compute the mean and median errors of them.
Specifically, errors are gauged through $L_2$ distances and we employ linear interpolation to extend the output of our model to the next 1s for evaluation.

\begin{table*}[t!]
	\caption{Evaluation results of motion prediction on nuScenes test set.  Our self-supervised methods, RigidFlow and RigidFlow++, outperform  self-supervised PillarMotion by a large margin.}
	\vspace{-3mm}
	\label{tab_nuscenes}
	\begin{center}
		\setlength{\tabcolsep}{4.5pt}
		\renewcommand\arraystretch{1.05}	
		\resizebox{2.0\columnwidth}{!}{
			\centering
			\begin{tabular}{l @{\hskip 0.4cm}|@{\hskip 0.2cm}c@{\hskip 0.2cm}| @{\hskip 0.3cm}c@{\hskip 0.3cm}  | c@{\hskip 0.4cm}c | c@{\hskip 0.4cm}c | c@{\hskip 0.4cm}c }
				\Xhline{1.4pt}
				\multirow{2}{*}{Method} & \multirow{2}{*}{Supervision} & \multirow{2}{*}{Modality} &  \multicolumn{2}{c|}{Static} & \multicolumn{2}{c|}{Speed $\leq$ 5m/s} &  \multicolumn{2}{c}{Speed $>$ 5m/s} \\  \cline{4-9}
				& & & Mean~$\downarrow$ & Median~$\downarrow$ & Mean~$\downarrow$ & Median~$\downarrow$ & Mean~$\downarrow$ & Median~$\downarrow$\\
				\Xhline{1.4pt}
				LSTM-ED \cite{schreiber2019long}&   Fully supervised& LiDAR & 0.0358 & 0 & 0.3551 & 0.1044 & 1.5885 & 1.0003 \\
				PillarMotion~\cite{luo2021self} & Fully supervised & LiDAR+Image & 0.0245 & 0 & 0.2286 & 0.0930 & 0.7784 & 0.4685\\
				MotionNet~\cite{wu2020motionnet}  & Fully supervised & LiDAR & 0.0201 & 0 & 0.2292 & 0.0952 & 0.9454 & 0.6180\\
				BE-STI~\cite{wang2022sti}  &  Fully supervised & LiDAR & 0.0220 & 0 & 0.2115 & 0.0929 & 0.7511 & 0.5413\\
				\hline
				WeakMotion~\cite{li2023weakly}&  Weakly supervised (100\% FG/BG masks) & LiDAR&   0.0243 &  0 &  0.3316 &  0.1201  &  1.6422 &  1.0319  \\
				WeakMotion~\cite{li2023weakly}&  Weakly supervised (0.1\% FG/BG masks) & LiDAR&   0.0426 &  0 &  0.4009 &  0.1195  &  2.1342 & 1.2061 \\
				\hline
				PillarMotion~\cite{luo2021self} & Self-supervised &  LiDAR+Image & 0.1620 &  0.0010 &  0.6972&  0.1758& 3.5504 & 2.0844\\
				\bf RigidFlow&  Self-supervised  &  LiDAR &  0.1090&  0&  0.3470& 0.1067 &  \bf 2.4117 & 1.3448\\
				\bf RigidFlow++& Self-supervised &  LiDAR & \bf 0.0580 &  \bf 0 & \bf 0.3097 & \bf 0.1001 & 2.4937& \bf 1.2662\\
				\Xhline{1.4pt}
			\end{tabular}
		}
	\end{center}
\end{table*}

\begin{figure*}[tb]
	\centering
	\includegraphics[height=6.8cm]{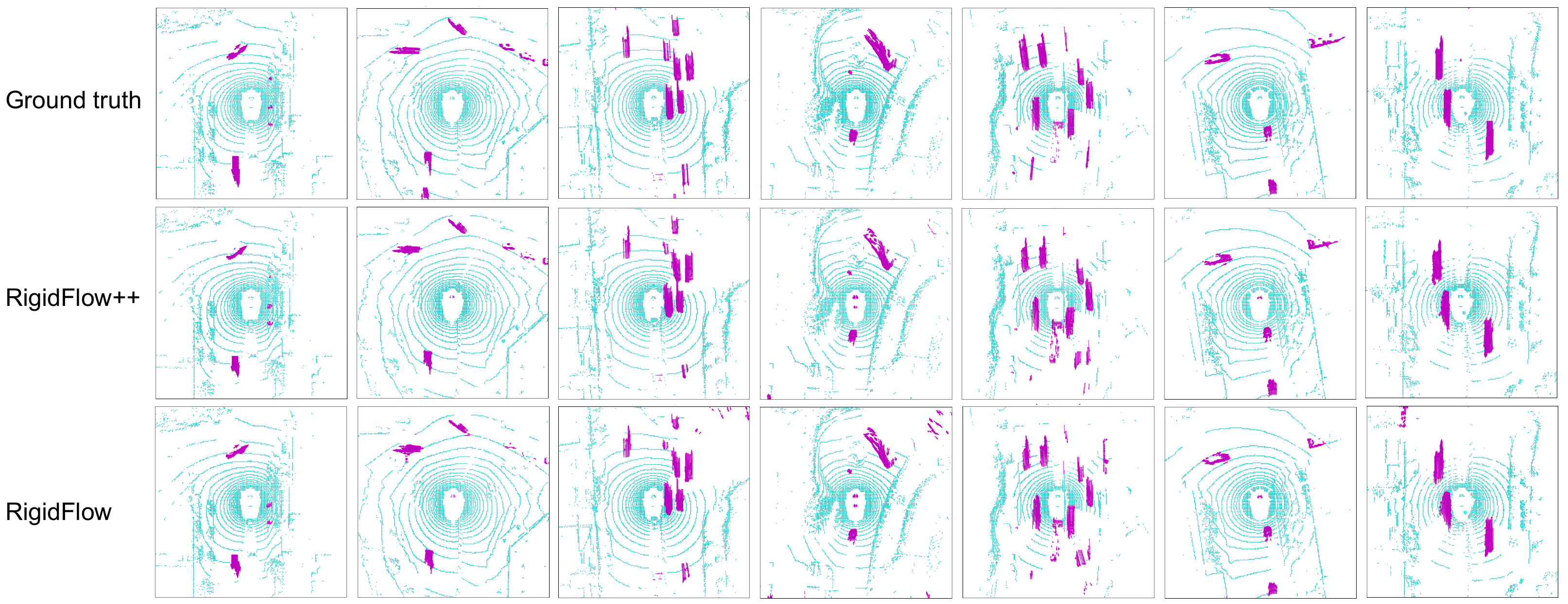}
	\vspace{-3mm}
	\caption{Qualitative results of motion prediction and static/moving segmentation on nuScenes. Top: Ground truth. Middle: Results of motion prediction model trained by our RigidFlow++. Bottom: Results of motion prediction model trained by our RigidFlow. We depict motion through an arrow connected to each cell and use different colors to indicate different motion states.
		{\textcolor[RGB]{255,0,255}{\textbf{Purple}}}: Moving objects;  	{\textcolor[RGB]{0,228,238}{\textbf{Cyan}}}: Static objects.}
	\label{fig_motion}	
\end{figure*}

\textbf{Comparison with state-of-the-art methods.}\quad
As presented in Table~\ref{tab_nuscenes}, without using any image information, our self-supervised methods, RigidFlow and RigidFlow++, outperform  self-supervised PillarMotion~\cite{luo2021self},  the state-of-the-art self-supervised method, by around 25\% on all evaluation metrics.
Specifically, our novel method, RigidFlow++, performs better than our RigidFlow on both static and slow groups.
Without using any manual annotations, our RigidFlow++ even outperforms weakly supervised WeakMotion~\cite{li2023weakly} on the slow speed group, which leverages foreground/background (FG/BG) binary masks as weak supervision.
The experimental results demonstrate the superiority of our method in self-supervised class-agnostic motion prediction.
Fig.~\ref{fig_motion} provides some qualitative results.

\section{Conclusion}

In this paper, we propose to produce pseudo scene flow labels by a piecewise  rigid motion estimation.
By decomposing the source point cloud into a set of local regions, we design a confidence-aware piecewise pseudo label generation module that alternately estimates point correspondences, confidence weights, and region-specific rigid transformations to generate reliable pseudo flow labels and their validity mask for self-supervised learning.
Comprehensive experiments on FlyingThings3D and KITTI datasets demonstrate that our proposed approach achieves state-of-the-art performance in self-supervised scene flow learning, without any ground truth scene flow for supervision, even outperforming some supervised counterparts.
In addition, our approach is further extended to the task of self-supervised class-agnostic motion prediction and achieves state-of-the-art performance on nuScenes dataset.

\bibliographystyle{ieeetr}
\bibliography{reference}

\end{document}